\documentclass[runningheads]{llncs}

\usepackage{eccv}

\usepackage{eccvabbrv}
\usepackage{verbatim}
\usepackage{graphicx}
\usepackage{booktabs}
\usepackage{kotex}
\usepackage{booktabs}
\usepackage{tabularx}
\usepackage{subcaption}
\usepackage{multirow}
\usepackage{bm}
\usepackage{cancel}
\usepackage{caption}
\usepackage{wrapfig}
\usepackage{comment}
\usepackage{lipsum}
\usepackage{listings}
\usepackage[accsupp]{axessibility}  
\newcommand{\ours}{PosterLlama}
\newcommand{\Fref}[1]{Figure~\ref{#1}}

\newcommand{\Sref}[1]{Section~\ref{#1}}
\newcommand{\Tref}[1]{Table~\ref{#1}}

\usepackage{hyperref}
\usepackage{orcidlink}

\begin{document}

\title{PosterLlama: Bridging Design Ability of Langauge Model to Content-Aware Layout Generation}%

\titlerunning{PosterLlama: Language Model to Content-Aware Layout Generation}

\author{Jaejung Seol\orcidlink{0009-0000-8751-6947} \and
Seojun Kim\orcidlink{0009-0008-8392-962X} \and
Jaejun Yoo\inst{*}\orcidlink{0000-0001-5252-9668}} 

\authorrunning{J. Seol et al.}

\institute{Laboratory of Advanced Imaging Technology (LAIT) \\
Ulsan National Institute of Science and Technology (UNIST)
\email{\{tjfwownd,seojun.kim,jaejun.yoo\}@unist.ac.kr}\\(*: corresponding author)}

\maketitle 

\begin{abstract}
Visual layout plays a critical role in graphic design fields such as advertising, posters, and web UI design.  
The recent trend toward content-aware layout generation through generative models has shown promise, yet it often overlooks the semantic intricacies of layout design by treating it as a simple numerical optimization. 
To bridge this gap, we introduce \ours, a network designed for generating visually and textually coherent layouts by reformatting layout elements into HTML code and leveraging the rich design knowledge within language models. 
Furthermore, we enhance the robustness of our model with a unique depth-based poster augmentation strategy. This ensures our generated layouts remain semantically rich but also visually appealing, even with limited data. 
Our extensive evaluations across several benchmarks demonstrate that \ours~outperforms existing methods in producing authentic and content-aware layouts. It supports an unparalleled range of conditions, including but not limited to content-aware layout generation, element conditional layout generation, and layout completion, among others, serving as a highly versatile user manipulation tool.
Project webpage: \href{https://lait-cvlab.github.io/PosterLlama}{PosterLlama}

\end{abstract}

\section{Introduction}
\label{sec:intro}
Layout is a fundamental element in graphic design, harmoniously arranging design elements such as logos and text to capture the reader's attention and effectively convey essential information. Its significance extends across various applications, ranging from graphic design, such as web UI, posters, and document typesetting, to downstream tasks like Human-Object Interaction(e.g., region-controlled image generation \cite{yang2023reco,li2023gligen} and layout-guided video generation\cite{lian2023llm}). Due to its versatile applications, layout design has garnered widespread attention by offering the potential to replace manual efforts for experts and achieve cost savings. Consequently, considerable research has been dedicated to creating controllable, high-quality layouts, aiming to enhance both aesthetic appeal and functional efficiency in design processes. \cite{li2019layoutgan,li2020attribute,yamaguchi2021canvasvae,jiang2022coarse,patil2020read,jyothi2019layoutvae,arroyo2021variational,chai2023layoutdm,inoue2023layoutdm,zhang2023layoutdiffusion,hui2023unifying,jiang2023layoutformer++,kong2022blt,gupta2021layouttransformer}

As the significance of visual content awareness in layout generation has gained increasing recognition, predicting layout structure that ensures text readability and maintains visual balance on the canvas has become more important. This is referred to as the poster layout generation problem.
Following the pioneering work of ContentGAN\cite{zheng2019content}, the first model to blend visual and textual data for generating layout structure, subsequent models have introduced sophisticated techniques to refine this process. CGL-GAN\cite{zhou2022composition} and DS-GAN\cite{hsu2023posterlayout} enhance spatial information encoding using multi-scale CNNs, employing transformers and CNN-LSTM architectures for decoding in layout generation, respectively. RADM\cite{fengheng2023relation} has expanded this paradigm by incorporating visual and textual content considerations into poster generation through diffusion models.

Despite advancements in layout generation, existing methods have limitations because they treat layout elements as simple numerical values for prediction, converting semantically rich categories into $0, 1$, and so on. This approach restricts the network's ability to learn beyond the distribution of layouts and fails to capture the semantic relationships among elements. For example, indicating that ``\texttt{The text} is placed over \texttt{the underlay}'' offers a more interpretable representation compared to stating ``\texttt{Element0} is placed over \texttt{Element1}''.
To tackle this challenge, recent efforts, such as LayoutPromtper\cite{lin2024layoutprompter}, Layout GPT\cite{feng2024layoutgpt}, and LayoutNUWA\cite{tang2023layoutnuwa}, have been focused on leveraging the powerful capabilities of language models. While these approaches show promise in producing high-quality layouts, they still face challenges in considering fine visual content. 

In this paper, we introduce PosterLlama, a novel model designed for generating poster layouts aware of both visual canvas and textual contents, integrating semantically rich layout elements. To achieve this, following the previous work~\cite{lin2024layoutprompter,tang2023layoutnuwa}, we reformat layout elements into HTML code to leverage the design knowledge embedded in language models. We additionally incorporate text descriptions into code format encouraging the model to learn text awareness.
To ensure the language model takes visual content into account, we design a two-stage training process inspired by the efficient vision-language training method\cite{zhu2023minigpt}. In the first stage, we train an adapter to connect the visual encoder with the LLM, and in the second stage, we train the model to generate HTML sequences. Given the challenges in assembling a large dataset due to copyright issues and other constraints with poster datasets, we propose a depth-based augmentation method that primarily focuses on the presence of salient objects within the poster. Lastly, to bridge the gap between layout generation tasks and real-world industrial applications, we propose a pipeline for generating advertisement posters that utilize a scene-text generation module. 

We conduct a comprehensive evaluation of PosterLlama using various methods. Experimental results demonstrate that PosterLlama is efficient with dataset size, showing state-of-the-art performance in nearly all metrics.
Notably, our model exhibits almost identical performance to real layouts regarding layout quality measurement, thanks to leveraging design knowledge from LLMs. To the best of our knowledge, PosterLlama is the first model that can handle all types of content-aware layout generation tasks, ranging from layout elements conditional generation to layout completion and refinement.
This capability suggests the applicability of PosterLlama to practical content-aware layout generation tasks.

\section{Related Work}
\subsubsection{Content-Agnostic Layout Generation}

Content-agnostic layout generation aims to create layouts without being restricted by specific content, like a canvas. Early works in layout generation \cite{cao2012automatic,schrier2008adaptive,kumar2011bricolage,o2014learning} rely on professionally designed templates or heuristic rules for generating layouts. These methods require task-specific professional knowledge and are often unable to generate a wide variety of layouts. To overcome these issues, a data-driven approach utilizing deep generative models is proposed. LayoutGAN \cite{li2019layoutgan} is the first approach introducing GAN to synthesize semantic and geometric layout elements. It adopted a differentiable wireframe rendering to incorporate visual attributes into the layout generation process. Subsequently, LayoutGAN has been enhanced to support attribute-conditioned design tasks\cite{li2020attribute}. 
Following the initial approaches, methods involving Variational Autoencoders\cite{yamaguchi2021canvasvae,jiang2022coarse,jyothi2019layoutvae,arroyo2021variational,patil2020read}, Diffusion models\cite{chai2023layoutdm,inoue2023layoutdm,zhang2023layoutdiffusion,hui2023unifying}, and transformers\cite{jiang2023layoutformer++,kong2022blt,gupta2021layouttransformer} have been introduced for both constrained and unconstrained layout generation tasks. Meanwhile, recent developments in LLM-based layout generation methods~\cite{feng2024layoutgpt,lin2024layoutprompter,tang2023layoutnuwa} have shown results that match those of traditional models. These advancements highlight the limitations of current layout generation approaches, specifically their insufficient understanding of semantic information among layout elements, as they mainly rely on numerical optimization methods. 
Among these methods, LayoutNUWA\cite{tang2023layoutnuwa} is the most closely related to ours, achieving state-of-the-art performance by fine-tuning LLMs to generate layouts in HTML format. However, unlike our approach, LayoutNUWA's scope is limited to content-agnostic layout generation. 

\subsubsection{Content-Aware Layout Generation}
With the advancement of content-agnostic layout generation methods, research incorporating content such as images and text into layout generation has received increased attention, particularly in generative\cite{zheng2019content,hsu2023posterlayout,fengheng2023relation,zhou2022composition,cao2022geometry} manner. ContentGAN\cite{zheng2019content} first considered the relationship not only between layout elements but also between layout and images. Although they benefit from content information, as they neglect composition like spatial information, they encounter the occlusion problem in poster layout generation. 
CGL-GAN\cite{zhou2022composition} and DS-GAN\cite{hsu2023posterlayout} employ an encoder-decoder architecture, utilizing a standard transformer and a CNN-LSTM for the decoder, respectively. By leveraging a ResNet-FPN network as the visual encoder and additionally injecting a saliency map, they demonstrate robust capabilities in generating a visual content-aware layout.  RADM\cite{fengheng2023relation} stands out as the first paper to integrate textual content, a highly significant attribute, into visual elements for layout generation, considering spatial composition. Through a refined design of the relation-aware module within the denoising diffusion model, RADM achieves realistic poster layout generation. Recently, leveraging the power of LLM, 
LayoutPrompter\cite{lin2024layoutprompter} has been introduced as a content-aware layout generation method. However, it has a limitation in achieving fine visual composition as it solely represents visual content information using the bounding box of salient objects. In contrast, our work takes into account visual encoding, enabling high-quality content-aware layout generation with a focus on fine visual composition.

\section{Method}
In this section, we introduce PosterLlama, an LLM-based multi-modal layout generation model. 
We redesign layout generation tasks as HTML sequence generation to leverage LLM's design knowledge effectively (\Sref{subsec:Layout Formation}). To incorporate visual understanding into LLM, we propose a two-stage training method for multi-modal layout generation (\Sref{subsec:Training Method}). We also develop a Layout Augmentation Module for enhanced layout robustness (\Sref{subsec:Depth-Guided Poster Augmentation}).
\subsection{Input Output Sequence Formatting}
\label{subsec:Input Output Sequence Formatting}
\begin{figure}[tb]
  \centering
  \includegraphics[width=\textwidth]{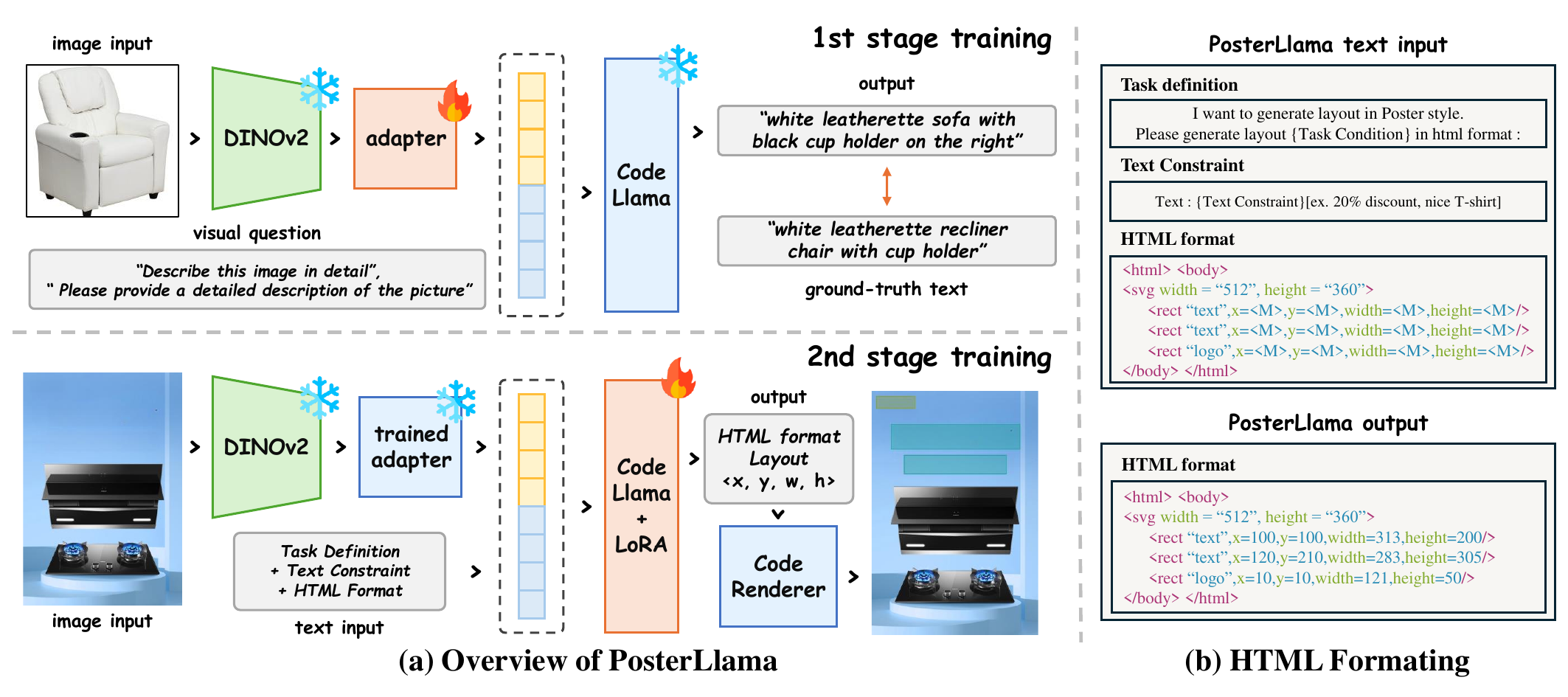}
  \caption{(a) The overall training step of PosterLlama. The first stage of training (top) involves training the adapter to learn visual alignment. In the second stage of training (bottom), we fine-tune the model for HTML formatted layout generation, utilizing the frozen adapter. (b) Illustration of input-output sequence in PosterLlama}
  \label{fig:posterllama}
\end{figure}
\subsubsection{Layout Formation}
\label{subsec:Layout Formation}

The goal of content-aware layout generation is to generate layouts constrained on given content condition $\mathcal{C}$. In our poster layout generation, the condition $\mathcal{C}$ is defined as multi-modal content (e.g., poster canvas, textual explanation). The layout is represented by a set of N elements $\{ e_i \}_{i=1}^N$, where $e_i=(t_i,s_i,c_i)$ consists of bounding box location $t_i = (x_i,y_i)$, size $s_i = (w_i,h_i)$, and category $c_i$. Following\cite{jiang2023layoutformer++}, in addition to conditions by content, the subset of layout elements can be also applied as a constraint. For example, Gen-IT~\cite{jiang2023layoutformer++} generates a layout constrained by element category types.
\subsubsection{HTML Formatting}
\label{subsec:HTML Formatting}

To leverage the extensive knowledge encapsulated in LLM for layout design, we represent layouts in the form of HTML sequence formations\cite{lin2024layoutprompter,tang2023layoutnuwa}. This method not only allows us to utilize design priors embedded in the LLM's training data, such as those from web-UI designs, but also offers a stronger expressive capability, compared to merely representing layout attributes as numerical values.
In line with the approach \cite{tang2023layoutnuwa}, we develop a template for generating text-aware layouts by constructing the model input sequence with a task definition, HTML formatting, and adding Text Constraints. The task definition, identified by \texttt{\{Task Condition\}}, specifies the conditions of the input sequence (for instance,  \texttt{\{``according to the categories and image''\}} in Gen-IT~\cite{jiang2023layoutformer++}). We employ HTML formatting to encapsulate the diverse tags characterizing the Web-UI layout, such as the \texttt{<rect>} tag, to wrap around the layout elements.
To facilitate conditional layout generation, we introduce a mask token \texttt{<M>}, prompting the LLM to predict this masked token. As we can readily imagine, layout elements do not inherently possess a specific order. However, organizing the input and output of the layout into 
a predefined order of mask tokens during the learning process with limited data and diverse conditions can lead to overfitting. Hence, we introduce a random permutation to the layout order while maintaining synchronization between input and output elements. Additionally, for efficient training with a focus on reducing the overall token length, We discretize the attributes of each element similarly to previous work \cite{chai2023layoutdm, zhang2023layoutdiffusion, inoue2023layoutdm}.
\subsection{Training Method}
\label{subsec:Training Method}
We employ LLM for the poster layout generation. To leverage the vision-language capabilities of the model for layout generation, we adopt a two-stage training approach inspired by the efficient Visual Question Answering training method of Mini-GPT4~\cite{zhu2023minigpt} with instruction tuning. In the first stage, we utilize a linear layer as an adapter to align the image encoder with the LLM, training only the adapter while keeping others frozen. This training is performed on an extensive collection of aligned image-text pairs. The encoded image feature of the encoder is encapsulated within \texttt{<img>} tokens and treated as a text token along with textual instructions:
\texttt{``<img><ImageFeature></img> Describe this image in detail.''}
For the visual encoder, we adopt DINOv2\cite{oquab2023dinov2}, drawing inspiration from recent advancements in visual embedding progress\cite{stein2024exposing}. In the second stage, while keeping the visual adapter frozen, we fine-tune the LLM to generate layouts using an HTML-formatted dataset described in \Sref{subsec:HTML Formatting}. 
To address the challenge of catastrophic forgetting and optimize the LLM fine-tuning process, our approach involves the utilization of LoRA as proposed by\cite{hu2021lora}. We train our model using cross-entropy loss as the objective function.

\subsection{Depth-Guided Poster Augmentation}
\label{subsec:Depth-Guided Poster Augmentation}

\begin{figure}[tb]
  \centering
  \includegraphics[width=\textwidth]{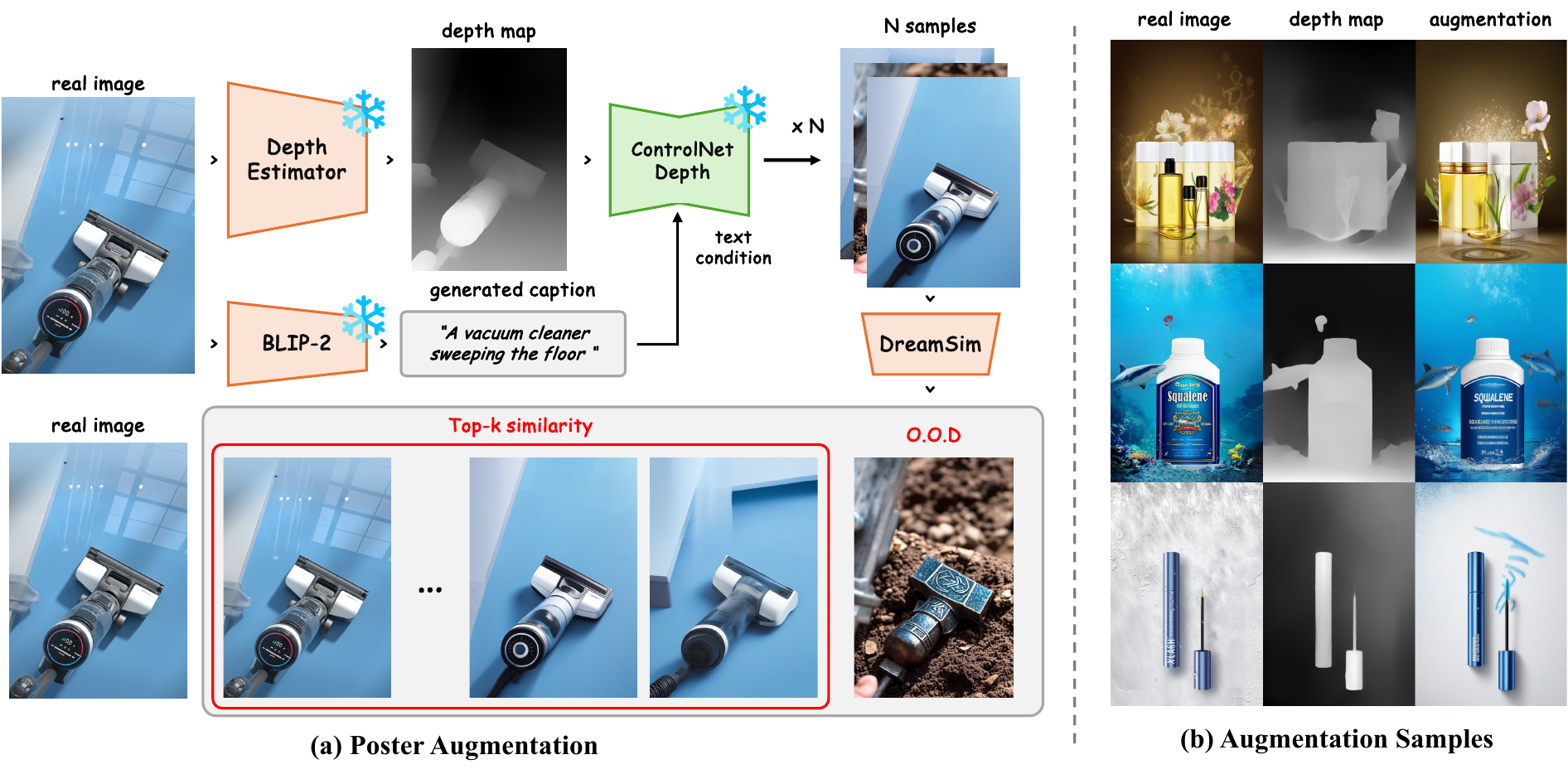}
  \caption{(a) The illustration of depth-guided poster augmentation. The estimated depth map and caption undergo multiple augmentations with the ControlNet, and the top-\textit{k} similarity samples with real images are selected as augmentation samples. (b) The illustration of augmented samples.
  }
  \label{fig:layoutaugmenatation}
\end{figure}
The performance of generative models fundamentally improves with diverse and rich data. However, poster datasets significantly lack the volume compared to the vast data pools, like LAION\cite{schuhmann2022laion} that are typically utilized for training foundational generative models. This scarcity, coupled with copyright issues related to the processed nature of poster images, makes the collection of large datasets challenging. 
To address the limitations, we introduce a novel poster augmentation illustrated in \Fref{fig:layoutaugmenatation}. It leverages depth-based augmentation, closely connected with salient objects\cite{zhao2022joint}, and a top-\textit{k} similarity selection in a two-step process, drawing inspiration from recent content-aware layout generation works that utilize salient object information as a visual condition~\cite{zhou2022composition,hsu2023posterlayout}.
In the depth-based augmentation phase, we employ ControlNet\cite{zhang2023adding}-Depth, a diffusion-based generative model conditioned on text and depth maps. To construct the text condition, we generate captions from original images using the high-quality captioning network, Blip-2, and formulate scripts like \texttt{``Please generate \{Caption\} in advertisement poster.''} Depth maps are estimated using readily available estimation networks. Despite ControlNet's ability to synthesize high-quality images, recent studies have shown that diffusion-generated images have undesirable artifacts~\cite{gandikota2023erasing}. 
Especially, artifacts that appear on the salient objects, can detrimentally affect the correlation between layout and image canvas, hindering the learning process of the network. To mitigate this problem, we introduce DreamSIM\cite{fu2023dreamsim}, a similarity measurement sensitive to layout and semantic content, to select the top-\textit{k} samples from $N$ generated samples. Here, we use $N=10$ and $\textit{k}=3$; \textit{i.e.,} per each image, three samples are selected for augmentation out of 10 generated samples.
The resultant images illustrate the generation of high-quality synthetic data with minimal changes, preserving the overall composition and salient objects. \Fref{fig:layoutaugmenatation}(b) shows augmented examples. 

\begin{table}[h]
  \centering
  \caption{Quantitative comparison with baselines on content-aware layout generation task. Text-aware Poster Llama is annotated as Poster Llama-T.}
  \begin{tabularx}{\textwidth}{@{}l *{4}{X} *{4}{X}@{}}
    \toprule
    \multirow{2}{*}{\texttt{Method}} & \multicolumn{6}{c}{\texttt{Graphic} } & \multicolumn{2}{c}{\texttt{Content}} \\
    \cmidrule(lr){2-7} \cmidrule(l){8-9}
    & \texttt{val}$\uparrow$ & \texttt{ove}$\downarrow$ & \texttt{ali}$\downarrow$ & \texttt{und\_l}$\uparrow$ & \texttt{und\_s}$\uparrow$ & \texttt{FD}$\downarrow$ & \texttt{rea}$\downarrow$ & \texttt{occ}$\downarrow$  \\
    \midrule
    \textbf{CGL-Dataset} & & & & & & & & \\
    Real-data           & 0.9839        & 0.0002            & 0.0017       & 0.9937    & 0.9884 & \ \ \ - & 0.2059 &  0.1399  \\
    DS GAN              & 0.8545        & 0.0379            & 0.0033        & 0.9193    & 0.6717 & 41.45 & 0.2121 & 0.1491  \\
    LayoutPrompter     & 0.9950        & 0.0035            & 0.0025        & 0.4473    & 0.3201 & 4.048 & 0.2314 & 0.3011  \\
    RADM                & 0.9962        & 0.0702            & 0.0008         & 0.9854     & 0.9277 & \bf0.729 & \bf0.1961 & \bf0.1391  \\
    Poster Llama (Ours)  & 0.9986        & 0.0052            & 0.0004      & \bf0.9982 & \bf0.9909 & 2.518 & 0.2072 & 0.1480  \\
    Poster Llama-T (Ours)& \bf0.9988   & \bf0.0024            & \bf0.0003     & 0.9918    & 0.9905 & 2.082 & 0.2042 & 0.1418 \\
    \bottomrule
    \textbf{PKU-Dataset} & & & & & & & & \\
    Real-data & 0.9997 & 0.0013 & 0.0021 & 0.9974 & 0.9909 &  \ \  \ -  & 0.1729 & 0.1828  \\
    DS GAN & 0.8740 & 0.0336 & 0.0037 & 0.8688 & 0.5746 & 39.96 & 0.2035 & 0.2256  \\
    LayoutPromtper & 0.9995 & 0.0036 & 0.0032 & 0.5789 & 0.4139 & \bf2.753 & 0.2124 & 0.2992  \\
    Poster Llama (Ours) & \bf1.000 & \bf0.0032 & \bf0.0009 & \bf0.9998& \bf0.9910  & 12.44 & \bf0.1875 & \bf0.2087  \\
    \hline
  \end{tabularx}
  \label{tab:performance_metrics}
\end{table}
\section{Experiments}

\subsection{Experimental setting}

\subsubsection{Datasets}

we utilize two publicly available datasets, CGL and PKU, which are poster data collected from e-commerce platforms. The PKU dataset consists of three elements - Logo, Text, and Underlay, while the CGL dataset includes  Embellishment as an additional element. CGL provides a total of 60,548 annotated pairs of poster-layout, along with 1,000 unannotated posters. On the other hand, the PKU dataset offers 9,974 annotated pairs and 905 unannotated posters. As CGL does not supply inpainted posters separately, similar to prior work \cite{zhou2022composition,fengheng2023relation}, we inpainted the dataset using an inpainting network. Additionally, because both CGL and PKU datasets do not provide text annotations, we opt for the CGL-v2 dataset to facilitate layout generation that incorporates both textual and visual content. Lastly, since PKU and CGL datasets do not provide annotated poster splits for validation and testing, we approximately divide the data into train/val/test sets in an 8:1:1.

\subsubsection{Baselines}

We compare our method with four Layout Generation studies. (1) DS-GAN\cite{hsu2023posterlayout} utilizes a CNN encoder and a CNN-LSTM decoder. Due to its specific element sorting algorithm, this model is unsuitable for constrained generation tasks. (2) LayoutPrompter\cite{lin2024layoutprompter} is a method for generating layouts through in-context learning without the need for training. Similar to our approach, it uses LLM for layout generation. While the original paper employed GPT-3 text-davinci-003, our study utilized the GPT-3.5 turbo instruct model, as the former is currently unavailable. (3) RADM\cite{fengheng2023relation} is a diffusion-based model capable of handling visual-textual content for layout generation. (4) RALF \cite{horita2024retrievalaugmented} is an autoregressive model that leverages a Transformer decoder and sophisticated retrieval augmentation. Since RALF uses a different dataset processing method from ours, we use the publicly released weights without retraining and evaluate on an augmented dataset at analysis which is a completely separate setting.

\subsubsection{Implementation details}
Our PosterLlama model is built upon a language model and utilizes the ViT\cite{dosovitskiy2020image} as its visual encoder. For the visual encoding process, we select the DINOv2-base\cite{oquab2023dinov2} model, which is capable of processing images with a resolution of 224$\times$224. The CodeLlaMA\cite{roziere2023code} model serves as the LLM component. The rationale behind choosing the visual encoder and LLM is thoroughly examined in the ablation study detailed in \Sref{subsec:Effect of Model Architecture}. For the first stage training, we use a combined 2M image-captioning dataset that includes images from LAION\cite{schuhmann2022laion}, Conceptual Captions\cite{sharma2018conceptual}, and SBU\cite{ordonez2011im2text}.   To train our model efficiently, we employ the DeepSpeed\cite{rasley2020deepspeed} Library, conducting the training for five days on two A100 GPUs, with a batch size of 32 and leveraging gradient accumulation techniques\cite{lamy2021layered}.  During inference, we utilize top-\textit{p} \cite{holtzman2019curious} sampling with a $p$ value of 0.9 and a sampling temperature of 0.7.
\subsection{Evaluation Metrics}
we follow the evaluation metrics in previous work \cite{hsu2023posterlayout,zhou2022composition,kikuchi2021constrained}, including two aspects: graphic measures, and composition-relevant measures.
\subsubsection{Graphic Measures} This measure assesses graphic quality by considering the relationships between layout elements, without taking the canvas into account. We evaluated six commonly used metrics for layout graphic measure: Validity(\texttt{val}),  Alignment (\texttt{ali}), Overlap (\texttt{ove}), Underlay (\texttt{und\_l}, \texttt{und\_s}), and Frechet Distance (\texttt{FD}). The validity is the ratio of valid elements greater than 0.1\% of the canvas. We calculate all of the measures with valid elements. \texttt{ali}, \texttt{ove}, and \texttt{und} are assessed based on the methodology outlined in \cite{hsu2023posterlayout}. Additionally, we compute the Frechet distance (\texttt{FD})\cite{heusel2017gans} in the feature space pretrained by \cite{kikuchi2021constrained}.
\subsubsection{Content Measures} The aim of content measures is to evaluate the harmony between the generated layout elements and the canvas, following appropriate design rules. We employ content measures such as Occlusion (\texttt{occ}), and Readability score (\texttt{rea}) for our assessment. Occlusion measures the extent of the area occluded between the well-defined layout and salient objects, with the intuition that a well-defined layout minimizes the extent to which it obscures salient objects. The Readability score evaluates the clarity of text elements by assessing their gradient variation in the image space.

\subsection{Comparison}
\subsubsection{Quantitative Result}

In this section, we assess the performance of our PosterLlama in comparison to DS-GAN\cite{hsu2023posterlayout}, LayoutPrompter\cite{lin2024layoutprompter}, and RADM\cite{fengheng2023relation}---all of which are sophisticatedly designed methods for layout generation. The evaluation is conducted using eight different metrics. Due to the absence of text annotations in the PKU dataset, we exclusively compare RADM's performance on the CGL dataset. The quantitative results of the annotated test split without user constraints are summarized in \Tref{tab:performance_metrics}. PosterLlama attains the highest scores across five metrics and the second-highest scores for \texttt{FD}, \texttt{rea}, and \texttt{occ} in the CGL dataset. Additionally, it achieves the highest score in all metrics except for \texttt{FD} in the PKU dataset. Further details on performance degradation will be discussed in the upcoming section addressing data leakage. 

\subsubsection{Data Leakage}
\label{subsec:Data leakage}
\begin{wrapfigure}{rb}{0.4\textwidth}
  \centering
  \includegraphics[width=0.38\textwidth]{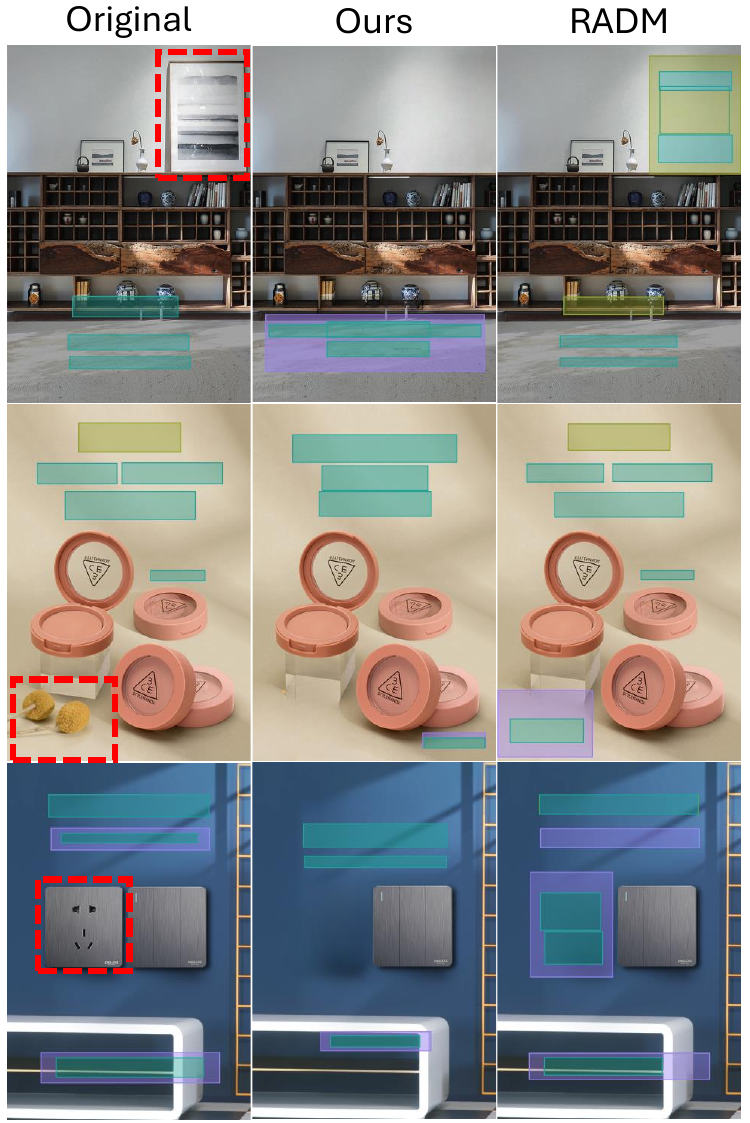}
  \caption{The data-leakage example of inpainted area. In addition to the original layout \textcolor{red}{Red} box area is deliberately erased and inpainted. Note that RADM puts the layouts precisely at the inpainted area.
  }
  \label{fig:radm-comparison}
\end{wrapfigure}

We note that PosterLlama's performance falls behind in three specific metrics (\texttt{FD}, \texttt{occ}, \texttt{rea}) compared to RADM. These three metrics only show high performance when the generated layout overlaps with the actual layout without introducing any novel elements. This is because they solely focus on the overlap with the canvas saliency map and the feature distance to the real layout, without evaluating the quality of the layout itself. Upon examining the layouts produced by RADM for test samples, we find that they always overlap with elements from the real layout. We hypothesize that this phenomenon may be due to artifacts introduced during the inpainting of the training data. Unlike PosterLlama, which uses a frozen visual encoder, RADM trains its visual encoder during the learning process, potentially allowing it to detect these artifacts due to data leakage. 
To test this hypothesis, we inpaint parts of objects in the test dataset and examine whether RADM's generated layouts overlap with these inpainted areas, as illustrated in \Fref{fig:radm-comparison}. Our results show that RADM places layout elements precisely at the inpainted areas (including the areas where original GT layouts were), suggesting it utilizes the inpainted artifacts to estimate the layout locations, unlike PosterLlama, which remain unaffected by the inpainted areas. Therefore, based on our analysis, we believe PosterLlama is more reliable than RADM despite its inferior performance in content metrics and \texttt{FD}. 

\subsubsection{Qualitative Result}
Here, we provide a qualitative comparison between our PosterLlama and the baseline method, as detailed in \Tref{tab:performance_metrics} and depicted in \Fref{fig:qualitative}. DS-GAN tends to generate layouts that are misaligned and overlapped, frequently positioning them in the top-left corner. This tendency is attributed to the fixed number of elements and the placement of non-element layouts at the top-left $(0,0,0,0)$ position. Although Layout Prompter yields highly aligned layouts, it lacks content awareness, leading to significant occlusion. On the other hand, RADM exhibits structures that closely resemble real data across all samples. In contrast, our PosterLlama approach demonstrates the ability to generate well-fitted and sensible layouts without overfitting the real data.
 
\subsection{Analysis}
\subsubsection{Effect of Model Architecture}
\label{subsec:Effect of Model Architecture}

\begin{table}[t]
  \centering
  \caption{Model architecture comparison. Initially based on Llama2+Eva-vit \cite{zhu2023minigpt}, our approach incorporates Dinov2 for visual encoding and CodeLlama tailored for effective code language processing. We denote CodeLlama as CL.} 
  \small 
  \begin{tabular}{lcccccccc}
    \toprule
    \normalsize PosterLlama & \texttt{\normalsize val}$\uparrow$ & \texttt{\normalsize ove}$\downarrow$ & \texttt{\normalsize ali}$\downarrow$ & \texttt{\normalsize und\_l}$\uparrow$ & \texttt{\normalsize und\_s}$\uparrow$ & \texttt{\normalsize FD}$\downarrow$ & \texttt{\normalsize rea}$\downarrow$ & \texttt{\normalsize occ}$\downarrow$ \\
    \midrule
    \normalsize Llama2+Eva-vit & 0.9974 & 0.0040 & 0.0009 & 0.9898 & 0.9820 & 7.07 & 0.2169 & 0.1741 \\
    \normalsize Llama2+DINOv2 & \textbf{0.9992} & 0.0042 & 0.0008 & 0.9682 & 0.9133 & 3.87 & 0.2111 & 0.1506 \\
    \normalsize CL+DINOv2 (Ours) & 0.9988 & \textbf{0.0024} & \textbf{0.0003} & \textbf{0.9918} & \textbf{0.9905} & \textbf{2.08} & \textbf{0.2042} & \textbf{0.1418} \\
    \bottomrule
  \end{tabular}
  \label{tab:Effects_of_Model_Architectures}
\end{table}

We delve into the architectural design choices for language models and vision transformers, as detailed in \Tref{tab:Effects_of_Model_Architectures}. Our initial approach leverages a language model for content-aware layout generation, primarily adopting the Llama2$+$Eva-vit 
configuration, as proposed by MiniGPT-v2\cite{chen2023minigpt}. We subsequently integrate the sophisticated visual encoder DINOv2, as detailed in \Sref{subsec:Training Method}, along with CodeLlama, a variant of Llama2 fine-tuned for code language processing. The integration of DINOv2, as evidenced in the Llama2 
results led to a significant improvement in the visual content awareness metric. Despite Llama2's adaptability to a broad spectrum of language formats, it is primarily not designed for code interpretation. Thus, switching to CodeLlama enables our model to achieve superior outcomes, demonstrating its efficacy in the context of content-aware layout generation. This adaptation underscores the importance of tailored model architecture in enhancing performance metrics, particularly in specialized applications such as visual content comprehension. 

\subsubsection{Effect of Augmentation}
We assess the impact of our depth-guided augmentation technique, as discussed in \Sref{subsec:Depth-Guided Poster Augmentation}. The detailed results are presented in \Tref{tab:overall_ablation_study}. The table indicates an overall positive effect of our augmentation method. In the data-rich CGL-Dataset, most metrics show performance improvements, except for \texttt{FD} and \texttt{und\_s}. The slight degradation in these metrics can be attributed to the increased diversity introduced by our augmentation method, impacting strict underlay \texttt{und\_s} and distribution \texttt{FD}. We additionally discuss this in supplementary material \texttt{B}.
Remarkably, our augmentation method demonstrates significant performance improvements in the PKU-Dataset, characterized by a limited number of data samples. It shows performance gains across all metrics, highlighting the effectiveness of our augmentation in scenarios with sparse data. This emphasizes the utility of our augmentation approach in enhancing model performance, particularly in data-scarce environments.

\begin{table}[t]
  \caption{Effect of depth-guided augmentation technique.}
  \begin{subtable}{\textwidth}
    \label{tab:augmentation_pku_part1}
    \centering
    \begin{tabularx}{\textwidth}{@{}l *{8}{>{\centering\arraybackslash}X}@{}}
      \toprule
              CGL-Dataset& \texttt{val}$\uparrow$     & \texttt{ove}$\downarrow$    & \texttt{ali}$\downarrow$     & \texttt{und\_l}$\uparrow$  & \texttt{und\_s}$\uparrow$     & \texttt{FD}$\downarrow$      & \texttt{rea}$\downarrow$     & \texttt{occ}$\downarrow$ \\
      \midrule
      Ours w/o aug & 0.9984  & 0.0029  &  0.0006  & 0.9903 & \bf0.9987   & \bf1.43  & 0.2058    & 0.1476 \\
      Ours & \bf0.9988 & \bf0.0024 & \bf0.0003 & \bf0.9918 & 0.9905        & 2.08 & \bf0.2042 &\bf0.1418 \\
      \bottomrule
    \end{tabularx}
  \end{subtable}

  \begin{subtable}{\textwidth}
    \label{tab:augmentation_pku_part2}
    \centering
    \begin{tabularx}{\textwidth}{@{}l *{9}{>{\centering\arraybackslash}X}@{}}
      \toprule
        PKU-Dataset& \texttt{val}$\uparrow$     & \texttt{ove}$\downarrow$    & \texttt{ali}$\downarrow$     & \texttt{und\_l}$\uparrow$  & \texttt{und\_s}$\uparrow$     &\texttt{FD}$\downarrow$     & \texttt{rea}$\downarrow$     &  \texttt{occ}$\downarrow$  \\
      \midrule
      Ours w/o aug  & \bf1.000    & 0.0087      &  0.0014     & 0.9911      & 0.9661      &   12.47     & 0.1882      & 0.2119   \\
      Ours          & \bf1.000    & \bf0.0032   & \bf0.0009   & \bf0.9986   & \bf0.9910   & \bf12.44 & \bf0.1875 & \bf0.2087 \\
      \bottomrule
    \end{tabularx}
  \end{subtable}
  \label{tab:overall_ablation_study}
\end{table}
\begin{table}[t]
\caption{Comparison on depth-guided augmented CGL-v2 test dataset.}
  \centering 
    \setcounter{table}{3}
  \begin{subtable}{\columnwidth} 
    \centering
    \begin{tabularx}{\textwidth}{@{}l *{4}{X} *{4}{X}@{}}
      \toprule
              Model & \texttt{val}$\uparrow$     & \texttt{ove}$\downarrow$    & \texttt{ali}$\downarrow$     & \texttt{und\_l}$\uparrow$  & \texttt{und\_s}$\uparrow$          & \texttt{rea}$\downarrow$     & \texttt{occ}$\downarrow$ \\
      \midrule
      DS-GAN            & 0.8451 & 0.0336 & 0.0039 & 0.8848 & 0.5969  & 0.1169 & 0.0597  \\

      RADM              & \bf1.0000 & 0.0079 & 0.0026 & 0.9029 & 0.6817  & \bf0.0973 & 0.0528  \\
      RALF              & \bf1.0000 & 0.0156 & 0.0044 & 0.9820 & 0.9666  & 0.1126 & 0.0595  \\
        PosterLlama-T   & \bf1.0000 & 0.0009 & 0.0028 & 0.9909 & 0.9883  & 0.1116 & 0.0575  \\
        PosterLlama     & \bf1.0000 & \bf0.0006 & \bf0.0001 & \bf1.0000 & \bf1.0000  & 0.1142 & \bf0.0513  \\
      \bottomrule
    \end{tabularx}
  \end{subtable}
  \label{tab:comparing with ralf}

\end{table}

\subsubsection{Evaluation on Augmented Samples}
We endeavor to collect an artifacts-free clean dataset; however, even if we create the dataset ourselves, assessment is unfeasible as the specifically trained textual encoder of RADM is not publicly available.  Therefore, for leakage-free evaluation, we evaluate models on CGL-v2 test samples using our augmentation technique, which effectively minimizes inpainting artifacts, as detailed in supplementary material \texttt{E.3}. Additionally, we evaluate the recent visual content-aware model, RALF\cite{horita2024retrievalaugmented} using the publically released model. To ensure a fair comparison, we also evaluate PosterLlama trained on the unaugmented dataset. The results demonstrate that the excessively high score of RADM at quantitative result decreases, indicating susceptibility to information leakage. In contrast, our model achieves the highest scores across most metrics, showcasing robustness to inpainting artifacts. 

\subsubsection{Conditional Generation}
\begin{figure}[tb]
  \centering
  \includegraphics[width=\linewidth]{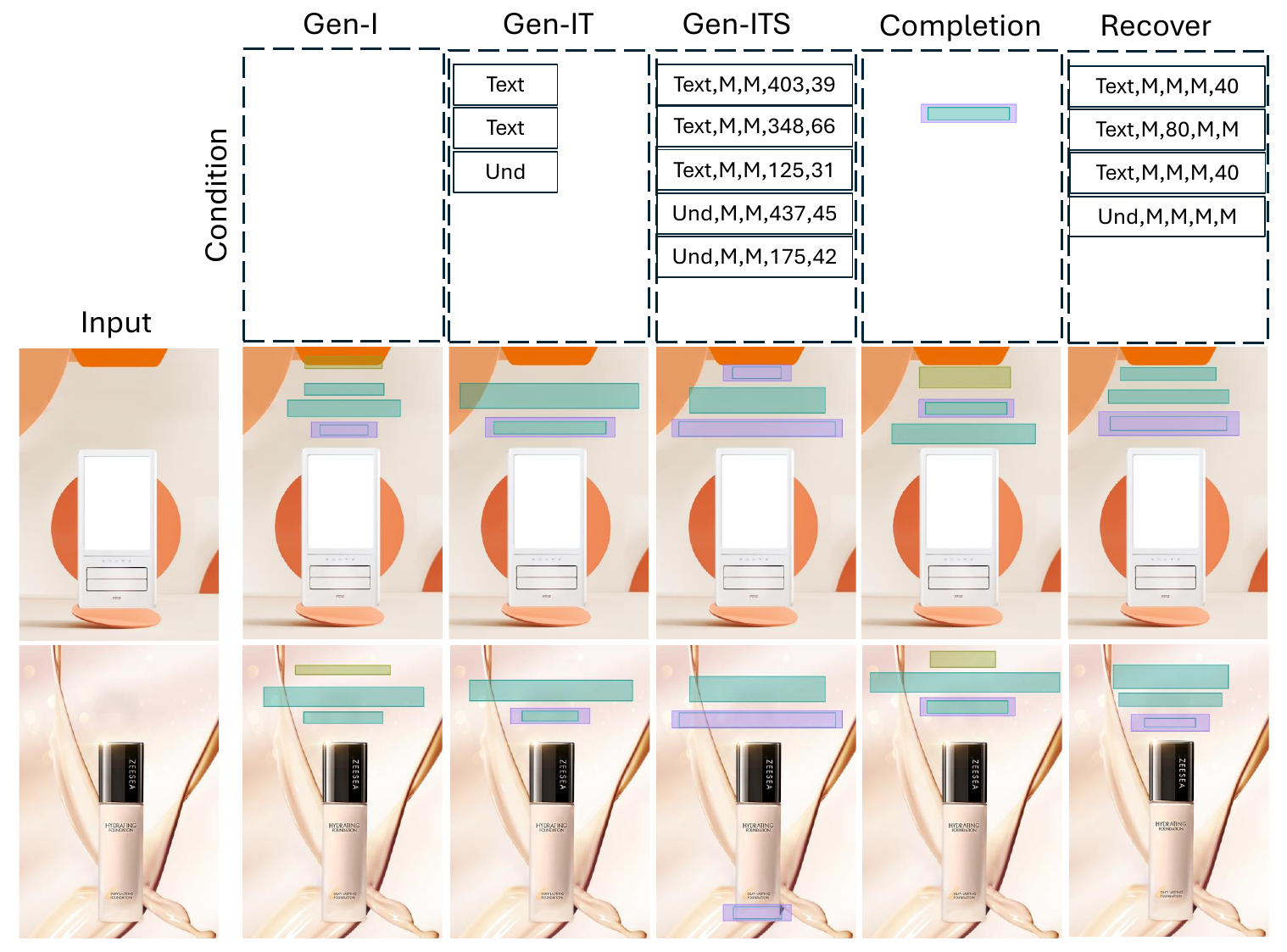}
  \caption{Visualization of PosterLlama across five conditional generations, with the conditions for layout generation positioned at the top. The order of elements is $(c, x, y, w, h)$. }
  \label{fig:Conditional}
\end{figure}
Our model is capable of achieving all types of user-conditioned generation defining the condition as text format, which is also a key feature of our method. We conduct experiments with 5 conditions as shown in \Fref{fig:Conditional}. The conditions are as follows: Gen-I involves image-conditioned layout generation, while Gen-IT and Gen-ITS additionally incorporate conditions based on category type and size. The completion task aims to generate a complete layout using partially placed elements. The recovery task involves restoring a randomly masked layout, with up to 80\% of elements being masked. As observed, PosterLlama is adept at handling a variety of user constraints while producing high-quality layouts. We construct these above 5 
conditions following\cite{jiang2023layoutformer++} and additional conditioning results are detailed in the supplementary materials.

\begin{figure}[tb]
  \centering
  \includegraphics[width=\textwidth]{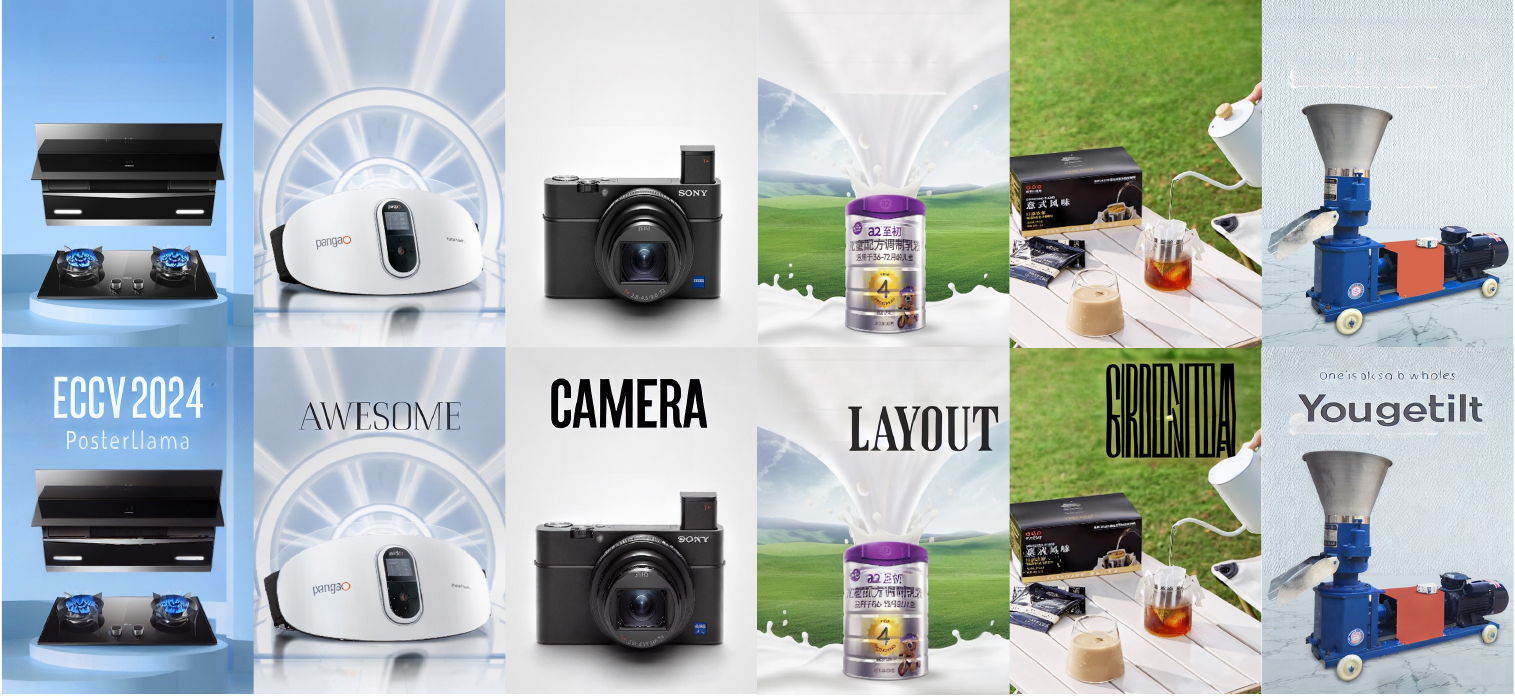}
  \caption{The example of proposed poster generation pipeline. The layout of scene text generation is generated by PosterLlama. Note that the right two samples exhibit unsatisfactory outcomes, attributed to inappropriate bounding box sizes.
  }
  \label{fig:poster-pipeline}
\end{figure}
\subsubsection{Poster Generation Pipeline}
To extend the applicability of layout generation to real-world industries, we propose a one-click advertisement poster generation pipeline. We utilize the scene-text generation model, i.e., AnyText\cite{tuo2023anytext} to generate a completed advertisement poster from the generated layout and background image. We first generate text layout from layout generation conditioning on the user text. Then we build a mask from the text layout. Lastly, we generate text-rendered images through the scene-text-generation model. By utilizing the proposed pipeline, users can easily get advertisement posters in just one click.  \Fref{fig:poster-pipeline} presents the outcomes of our poster generation pipelines. As demonstrated, our poster generation pipeline showcases applicability in creating plausible advertisement posters, indicating its usability in practical settings.

\section{Limitation}
We acknowledge two limitations as follows. Firstly, to accommodate the code language format suitable for CodeLlama, we translate the Chinese words provided in CGL-v2 into English for application in CodeLlama.  Consequently, in our text-aware layout generation scheme, it becomes challenging to fully consider the text length in the generated layout. Furthermore, recent studies in scene text generation have highlighted issues with degraded inpainted text quality when there is a mismatch between the sizes of the text and its bounding box \cite{liu2022character}. When we apply this scene text generation to our proposed poster generation pipeline, as shown in \Tref{fig:poster-pipeline} right two samples, we observe occurrences where the quality of the generated poster is compromised due to discrepancies in text length. Secondly, PosterLlama's reliance on the LLM foundation model makes it less scalable compared to other models. This limitation poses challenges when attempting to deploy PosterLlama on edge devices.
\section{Conclusion}
In this paper, we present PosterLlama, both a visual and largely unexplored textual content-aware layout generation method utilizing LLM. For content-aware layout generation, we leverage an efficient Visual Question-answering training method to instill visual awareness into the LLM, handling the layout in a code format suitable for the language model. To overcome data scarcity, we propose depth-guided augmentation using off-the-shelf generative models, which also alleviates inpainting artifacts enabling a fair evaluation. Our extensive experiments demonstrate that PosterLlama outperforms existing approaches, achieving diverse conditional generation by handling conditions in text format and being robust against learning shortcuts caused by inpainting artifacts. Thanks to this robustness and our augmentation method, we demonstrate that PosterLlama is highly effective with small datasets and adaptable for real-world applications.

\begin{figure}[t]
  \centering
  \includegraphics[width=\textwidth]{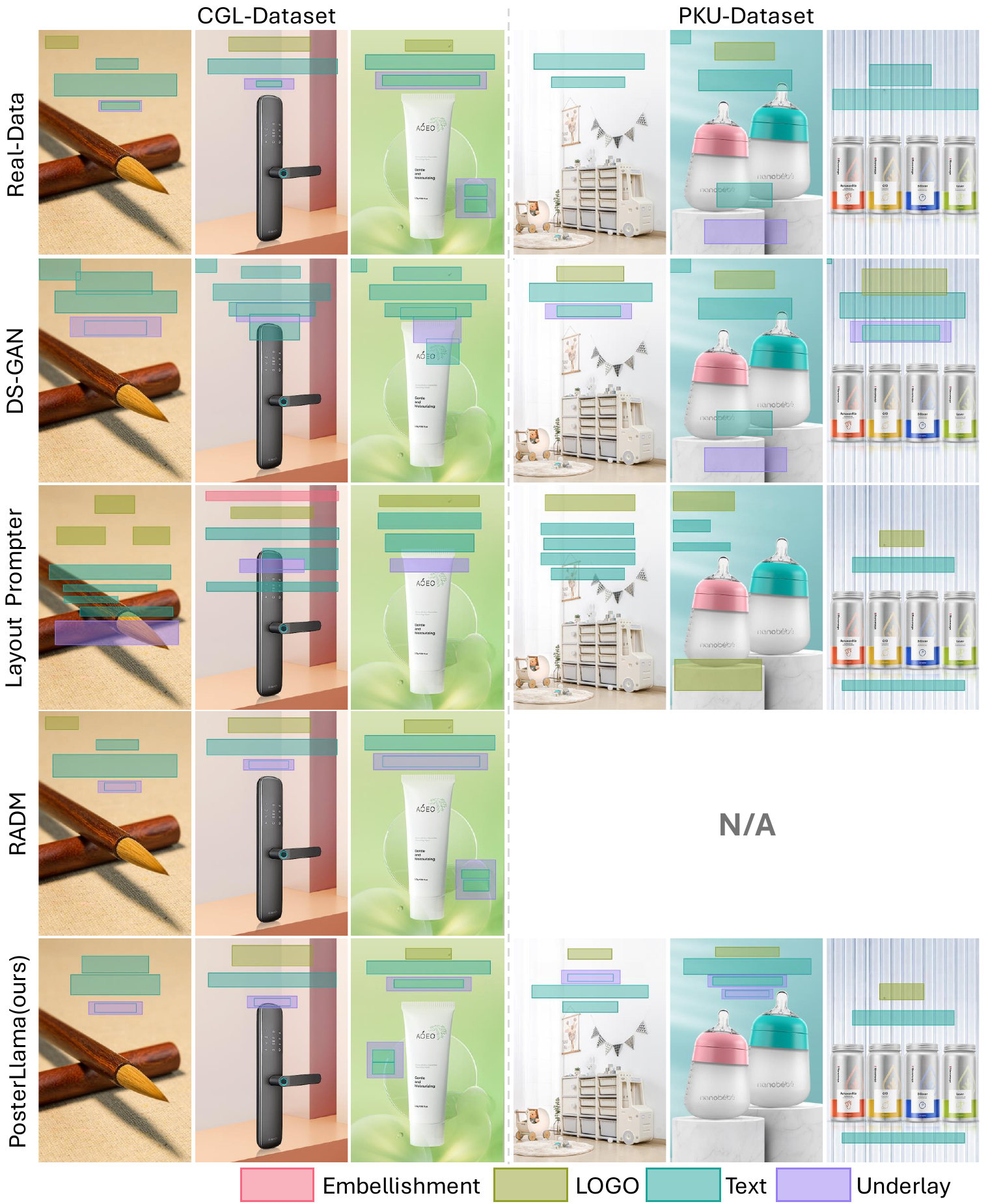}
  \caption{Qualitative comparison of image conditioned generation with baseline models about PKU and CGL-Dataset. Each comparison is implemented on annotated test samples. PKU Dataset lacks text descriptions, hence RADM does not proceed.
  }
  \label{fig:qualitative}
\end{figure}

\clearpage
\subsubsection{Acknowledgements} 
This work was supported by the National Research Foundation of Korea (NRF) grant funded by the Korea government (MSIT) (No.2022\ R1C1C1008496), Institute of Information \& communications Technology Planning \& Evaluation(IITP) grant funded by the Korea government(MSIT)(No.RS-2020-II201336, Artificial Intelligence graduate school support(UNIST), RS-2022-II220959, (Part 2) Few-Shot Learning of Causal Inference in Vision and Language for Decision Making, RS-2021-II212068, Comprehensive Video Understanding and Generation with Knowledge-based Deep Logic Neural Network), and the Ministry of Science and ICT(MSIT, Korea) \& Gwangju Metropolitan City (Artificial intelligence industrial convergence cluster development project). We also thank the supercomputing resources of the UNIST Supercomputing Center and Byung-hak Kim from the CJ AI Center for valuable discussions.

\bibliographystyle{splncs04}
\bibliography{egbib}

\begin{thebibliography}{10}
\providecommand{\url}[1]{\texttt{#1}}
\providecommand{\urlprefix}{URL }
\providecommand{\doi}[1]{https://doi.org/#1}

\bibitem{arroyo2021variational}
Arroyo, D.M., Postels, J., Tombari, F.: Variational transformer networks for layout generation. In: Proceedings of the IEEE/CVF Conference on Computer Vision and Pattern Recognition. pp. 13642--13652 (2021)

\bibitem{cao2012automatic}
Cao, Y., Chan, A.B., Lau, R.W.: Automatic stylistic manga layout. ACM Transactions on Graphics (TOG)  \textbf{31}(6),  1--10 (2012)

\bibitem{cao2022geometry}
Cao, Y., Ma, Y., Zhou, M., Liu, C., Xie, H., Ge, T., Jiang, Y.: Geometry aligned variational transformer for image-conditioned layout generation. In: Proceedings of the 30th ACM International Conference on Multimedia. pp. 1561--1571 (2022)

\bibitem{chai2023layoutdm}
Chai, S., Zhuang, L., Yan, F.: Layoutdm: Transformer-based diffusion model for layout generation. In: Proceedings of the IEEE/CVF Conference on Computer Vision and Pattern Recognition. pp. 18349--18358 (2023)

\bibitem{chen2023minigpt}
Chen, J., Zhu, D., Shen, X., Li, X., Liu, Z., Zhang, P., Krishnamoorthi, R., Chandra, V., Xiong, Y., Elhoseiny, M.: Minigpt-v2: large language model as a unified interface for vision-language multi-task learning. arXiv preprint arXiv:2310.09478  (2023)

\bibitem{dosovitskiy2020image}
Dosovitskiy, A., Beyer, L., Kolesnikov, A., Weissenborn, D., Zhai, X., Unterthiner, T., Dehghani, M., Minderer, M., Heigold, G., Gelly, S., et~al.: An image is worth 16x16 words: Transformers for image recognition at scale. arXiv preprint arXiv:2010.11929  (2020)

\bibitem{feng2024layoutgpt}
Feng, W., Zhu, W., Fu, T.j., Jampani, V., Akula, A., He, X., Basu, S., Wang, X.E., Wang, W.Y.: Layoutgpt: Compositional visual planning and generation with large language models. Advances in Neural Information Processing Systems  \textbf{36} (2023)

\bibitem{fu2023dreamsim}
Fu, S., Tamir, N., Sundaram, S., Chai, L., Zhang, R., Dekel, T., Isola, P.: Dreamsim: Learning new dimensions of human visual similarity using synthetic data. arXiv preprint arXiv:2306.09344  (2023)

\bibitem{gandikota2023erasing}
Gandikota, R., Materzynska, J., Fiotto-Kaufman, J., Bau, D.: Erasing concepts from diffusion models. arXiv preprint arXiv:2303.07345  (2023)

\bibitem{gupta2021layouttransformer}
Gupta, K., Lazarow, J., Achille, A., Davis, L.S., Mahadevan, V., Shrivastava, A.: Layouttransformer: Layout generation and completion with self-attention. In: Proceedings of the IEEE/CVF International Conference on Computer Vision. pp. 1004--1014 (2021)

\bibitem{heusel2017gans}
Heusel, M., Ramsauer, H., Unterthiner, T., Nessler, B., Hochreiter, S.: Gans trained by a two time-scale update rule converge to a local nash equilibrium. Advances in neural information processing systems  \textbf{30} (2017)

\bibitem{holtzman2019curious}
Holtzman, A., Buys, J., Du, L., Forbes, M., Choi, Y.: The curious case of neural text degeneration. arXiv preprint arXiv:1904.09751  (2019)

\bibitem{horita2024retrievalaugmented}
Horita, D., Inoue, N., Kikuchi, K., Yamaguchi, K., Aizawa, K.: {Retrieval-Augmented Layout Transformer for Content-Aware Layout Generation}. In: CVPR (2024)

\bibitem{hsu2023posterlayout}
Hsu, H.Y., He, X., Peng, Y., Kong, H., Zhang, Q.: Posterlayout: A new benchmark and approach for content-aware visual-textual presentation layout. In: Proceedings of the IEEE/CVF Conference on Computer Vision and Pattern Recognition. pp. 6018--6026 (2023)

\bibitem{hu2021lora}
Hu, E.J., Shen, Y., Wallis, P., Allen-Zhu, Z., Li, Y., Wang, S., Wang, L., Chen, W.: Lora: Low-rank adaptation of large language models. arXiv preprint arXiv:2106.09685  (2021)

\bibitem{hui2023unifying}
Hui, M., Zhang, Z., Zhang, X., Xie, W., Wang, Y., Lu, Y.: Unifying layout generation with a decoupled diffusion model. In: Proceedings of the IEEE/CVF Conference on Computer Vision and Pattern Recognition. pp. 1942--1951 (2023)

\bibitem{inoue2023layoutdm}
Inoue, N., Kikuchi, K., Simo-Serra, E., Otani, M., Yamaguchi, K.: Layoutdm: Discrete diffusion model for controllable layout generation. In: Proceedings of the IEEE/CVF Conference on Computer Vision and Pattern Recognition. pp. 10167--10176 (2023)

\bibitem{jiang2023layoutformer++}
Jiang, Z., Guo, J., Sun, S., Deng, H., Wu, Z., Mijovic, V., Yang, Z.J., Lou, J.G., Zhang, D.: Layoutformer++: Conditional graphic layout generation via constraint serialization and decoding space restriction. In: Proceedings of the IEEE/CVF Conference on Computer Vision and Pattern Recognition. pp. 18403--18412 (2023)

\bibitem{jiang2022coarse}
Jiang, Z., Sun, S., Zhu, J., Lou, J.G., Zhang, D.: Coarse-to-fine generative modeling for graphic layouts. In: Proceedings of the AAAI conference on artificial intelligence. vol.~36, pp. 1096--1103 (2022)

\bibitem{jyothi2019layoutvae}
Jyothi, A.A., Durand, T., He, J., Sigal, L., Mori, G.: Layoutvae: Stochastic scene layout generation from a label set. In: Proceedings of the IEEE/CVF International Conference on Computer Vision. pp. 9895--9904 (2019)

\bibitem{kikuchi2021constrained}
Kikuchi, K., Simo-Serra, E., Otani, M., Yamaguchi, K.: Constrained graphic layout generation via latent optimization. In: Proceedings of the 29th ACM International Conference on Multimedia. pp. 88--96 (2021)

\bibitem{kong2022blt}
Kong, X., Jiang, L., Chang, H., Zhang, H., Hao, Y., Gong, H., Essa, I.: Blt: bidirectional layout transformer for controllable layout generation. In: European Conference on Computer Vision. pp. 474--490. Springer (2022)

\bibitem{kumar2011bricolage}
Kumar, R., Talton, J.O., Ahmad, S., Klemmer, S.R.: Bricolage: example-based retargeting for web design. In: Proceedings of the SIGCHI Conference on Human Factors in Computing Systems. pp. 2197--2206 (2011)

\bibitem{lamy2021layered}
Lamy-Poirier, J.: Layered gradient accumulation and modular pipeline parallelism: fast and efficient training of large language models. arXiv preprint arXiv:2106.02679  (2021)

\bibitem{fengheng2023relation}
Li, F., Liu, A., Feng, W., Zhu, H., Li, Y., Zhang, Z., Lv, J., Zhu, X., Shen, J., Lin, Z., Shao, J.: Relation-aware diffusion model for controllable poster layout generation. In: Proceedings of the 32nd ACM International Conference on Information and Knowledge Management. p. 1249–1258 (2023)

\bibitem{li2019layoutgan}
Li, J., Yang, J., Hertzmann, A., Zhang, J., Xu, T.: Layoutgan: Generating graphic layouts with wireframe discriminators. arXiv preprint arXiv:1901.06767  (2019)

\bibitem{li2020attribute}
Li, J., Yang, J., Zhang, J., Liu, C., Wang, C., Xu, T.: Attribute-conditioned layout gan for automatic graphic design. IEEE Transactions on Visualization and Computer Graphics  \textbf{27}(10),  4039--4048 (2020)

\bibitem{li2023gligen}
Li, Y., Liu, H., Wu, Q., Mu, F., Yang, J., Gao, J., Li, C., Lee, Y.J.: Gligen: Open-set grounded text-to-image generation. In: Proceedings of the IEEE/CVF Conference on Computer Vision and Pattern Recognition. pp. 22511--22521 (2023)

\bibitem{lian2023llm}
Lian, L., Shi, B., Yala, A., Darrell, T., Li, B.: Llm-grounded video diffusion models. arXiv preprint arXiv:2309.17444  (2023)

\bibitem{lin2024layoutprompter}
Lin, J., Guo, J., Sun, S., Yang, Z., Lou, J.G., Zhang, D.: Layoutprompter: Awaken the design ability of large language models. Advances in Neural Information Processing Systems  \textbf{36} (2023)

\bibitem{liu2022character}
Liu, R., Garrette, D., Saharia, C., Chan, W., Roberts, A., Narang, S., Blok, I., Mical, R., Norouzi, M., Constant, N.: Character-aware models improve visual text rendering. arXiv preprint arXiv:2212.10562  (2022)

\bibitem{oquab2023dinov2}
Oquab, M., Darcet, T., Moutakanni, T., Vo, H., Szafraniec, M., Khalidov, V., Fernandez, P., Haziza, D., Massa, F., El-Nouby, A., et~al.: Dinov2: Learning robust visual features without supervision. arXiv preprint arXiv:2304.07193  (2023)

\bibitem{ordonez2011im2text}
Ordonez, V., Kulkarni, G., Berg, T.: Im2text: Describing images using 1 million captioned photographs. Advances in neural information processing systems  \textbf{24} (2011)

\bibitem{o2014learning}
O’Donovan, P., Agarwala, A., Hertzmann, A.: Learning layouts for single-pagegraphic designs. IEEE transactions on visualization and computer graphics  \textbf{20}(8),  1200--1213 (2014)

\bibitem{patil2020read}
Patil, A.G., Ben-Eliezer, O., Perel, O., Averbuch-Elor, H.: Read: Recursive autoencoders for document layout generation. In: Proceedings of the IEEE/CVF Conference on Computer Vision and Pattern Recognition Workshops. pp. 544--545 (2020)

\bibitem{rahman2021ruite}
Rahman, S., Sermuga~Pandian, V.P., Jarke, M.: Ruite: Refining ui layout aesthetics using transformer encoder. In: 26th International Conference on Intelligent User Interfaces-Companion. pp. 81--83 (2021)

\bibitem{rasley2020deepspeed}
Rasley, J., Rajbhandari, S., Ruwase, O., He, Y.: Deepspeed: System optimizations enable training deep learning models with over 100 billion parameters. In: Proceedings of the 26th ACM SIGKDD International Conference on Knowledge Discovery \& Data Mining. pp. 3505--3506 (2020)

\bibitem{roziere2023code}
Roziere, B., Gehring, J., Gloeckle, F., Sootla, S., Gat, I., Tan, X.E., Adi, Y., Liu, J., Remez, T., Rapin, J., et~al.: Code llama: Open foundation models for code. arXiv preprint arXiv:2308.12950  (2023)

\bibitem{schrier2008adaptive}
Schrier, E., Dontcheva, M., Jacobs, C., Wade, G., Salesin, D.: Adaptive layout for dynamically aggregated documents. In: Proceedings of the 13th international conference on Intelligent user interfaces. pp. 99--108 (2008)

\bibitem{schuhmann2022laion}
Schuhmann, C., Beaumont, R., Vencu, R., Gordon, C., Wightman, R., Cherti, M., Coombes, T., Katta, A., Mullis, C., Wortsman, M., et~al.: Laion-5b: An open large-scale dataset for training next generation image-text models. Advances in Neural Information Processing Systems  \textbf{35},  25278--25294 (2022)

\bibitem{sharma2018conceptual}
Sharma, P., Ding, N., Goodman, S., Soricut, R.: Conceptual captions: A cleaned, hypernymed, image alt-text dataset for automatic image captioning. In: Proceedings of the 56th Annual Meeting of the Association for Computational Linguistics (Volume 1: Long Papers). pp. 2556--2565 (2018)

\bibitem{stein2024exposing}
Stein, G., Cresswell, J., Hosseinzadeh, R., Sui, Y., Ross, B., Villecroze, V., Liu, Z., Caterini, A.L., Taylor, E., Loaiza-Ganem, G.: Exposing flaws of generative model evaluation metrics and their unfair treatment of diffusion models. Advances in Neural Information Processing Systems  \textbf{36} (2024)

\bibitem{tang2023layoutnuwa}
Tang, Z., Wu, C., Li, J., Duan, N.: Layoutnuwa: Revealing the hidden layout expertise of large language models. arXiv preprint arXiv:2309.09506  (2023)

\bibitem{tuo2023anytext}
Tuo, Y., Xiang, W., He, J.Y., Geng, Y., Xie, X.: Anytext: Multilingual visual text generation and editing. arXiv preprint arXiv:2311.03054  (2023)

\bibitem{yamaguchi2021canvasvae}
Yamaguchi, K.: Canvasvae: Learning to generate vector graphic documents. In: Proceedings of the IEEE/CVF International Conference on Computer Vision. pp. 5481--5489 (2021)

\bibitem{yang2023reco}
Yang, Z., Wang, J., Gan, Z., Li, L., Lin, K., Wu, C., Duan, N., Liu, Z., Liu, C., Zeng, M., et~al.: Reco: Region-controlled text-to-image generation. In: Proceedings of the IEEE/CVF Conference on Computer Vision and Pattern Recognition. pp. 14246--14255 (2023)

\bibitem{zhang2023layoutdiffusion}
Zhang, J., Guo, J., Sun, S., Lou, J.G., Zhang, D.: {LayoutDiffusion: Improving Graphic Layout Generation by Discrete Diffusion Probabilistic Models}. In: Proceedings of the IEEE/CVF International Conference on Computer Vision (ICCV) (2023)

\bibitem{zhang2023adding}
Zhang, L., Rao, A., Agrawala, M.: Adding conditional control to text-to-image diffusion models. In: Proceedings of the IEEE/CVF International Conference on Computer Vision. pp. 3836--3847 (2023)

\bibitem{zhao2022joint}
Zhao, X., Pang, Y., Zhang, L., Lu, H.: Joint learning of salient object detection, depth estimation and contour extraction. IEEE Transactions on Image Processing  \textbf{31},  7350--7362 (2022)

\bibitem{zheng2019content}
Zheng, X., Qiao, X., Cao, Y., Lau, R.W.: Content-aware generative modeling of graphic design layouts. ACM Transactions on Graphics (TOG)  \textbf{38}(4),  1--15 (2019)

\bibitem{zhou2022composition}
Zhou, M., Xu, C., Ma, Y., Ge, T., Jiang, Y., Xu, W.: Composition-aware graphic layout gan for visual-textual presentation designs. arXiv preprint arXiv:2205.00303  (2022)

\bibitem{zhu2023minigpt}
Zhu, D., Chen, J., Shen, X., Li, X., Elhoseiny, M.: Minigpt-4: Enhancing vision-language understanding with advanced large language models. arXiv preprint arXiv:2304.10592  (2023)

\end{thebibliography}

\clearpage
\appendix

\section{Code Availability}
Our code is available at \href{https://lait-cvlab.github.io/PosterLlama}{https://lait-cvlab.github.io/PosterLlama}
\section{Effect of Data Size}
\begin{figure}[t]
    \centering
    \begin{tabular}{ccc}
        \begin{subfigure}{0.5\textwidth}
            \centering
            \includegraphics[width=\textwidth]{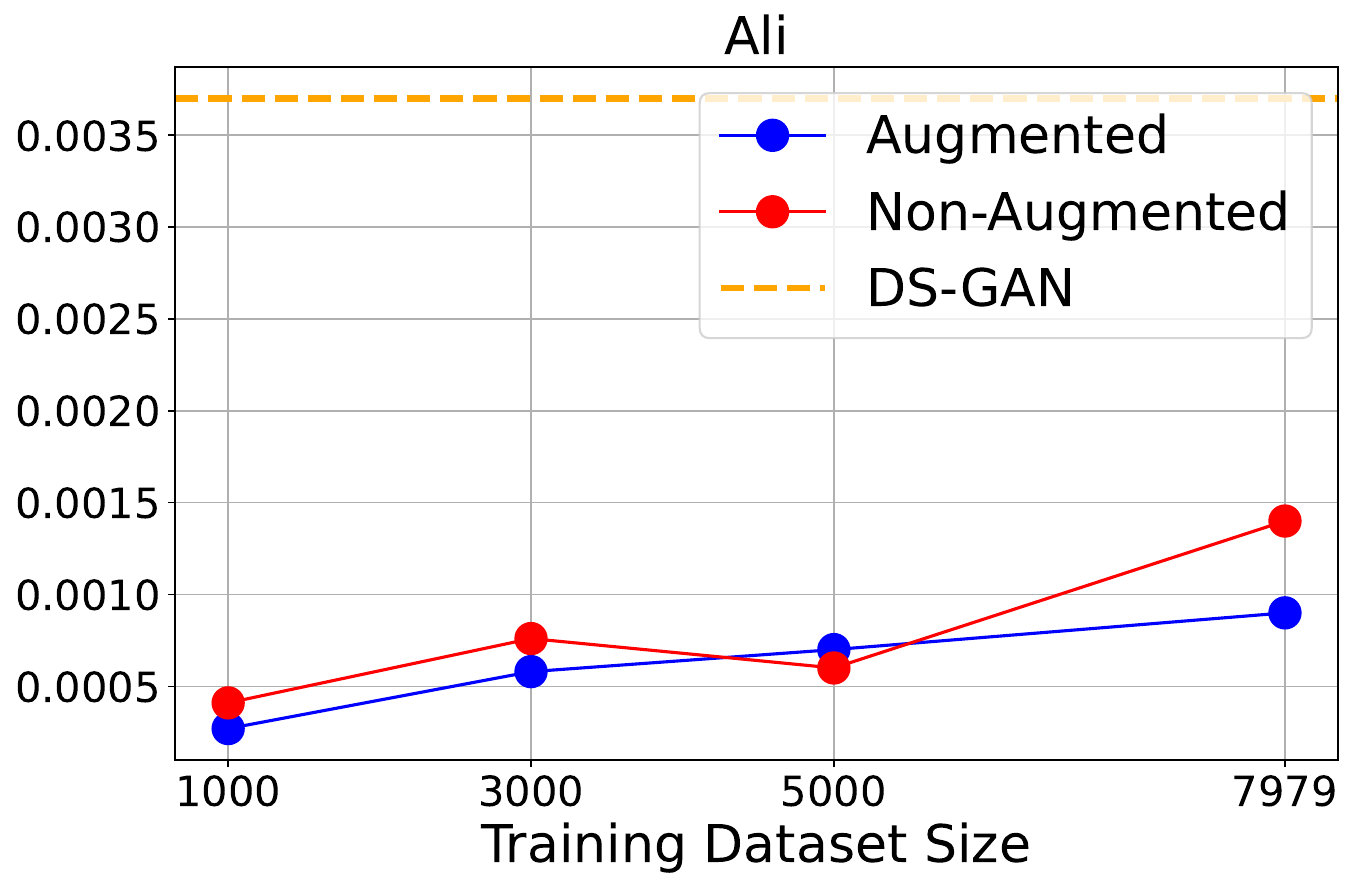}
        \end{subfigure} &
        \begin{subfigure}{0.465\textwidth}
            \centering
            \includegraphics[width=\textwidth]{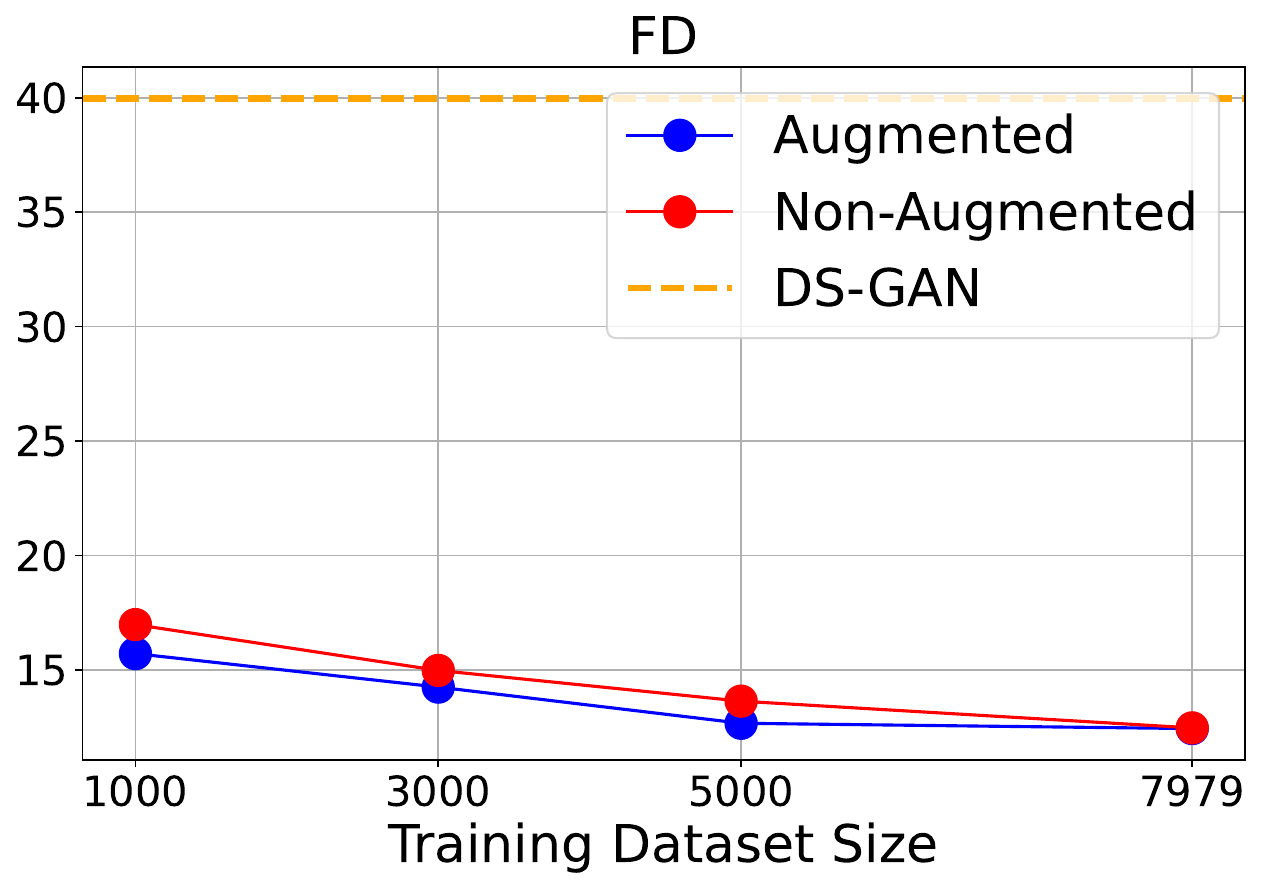}
        \end{subfigure} & \\
        \begin{subfigure}{0.48\textwidth}
            \centering
            \includegraphics[width=\textwidth]{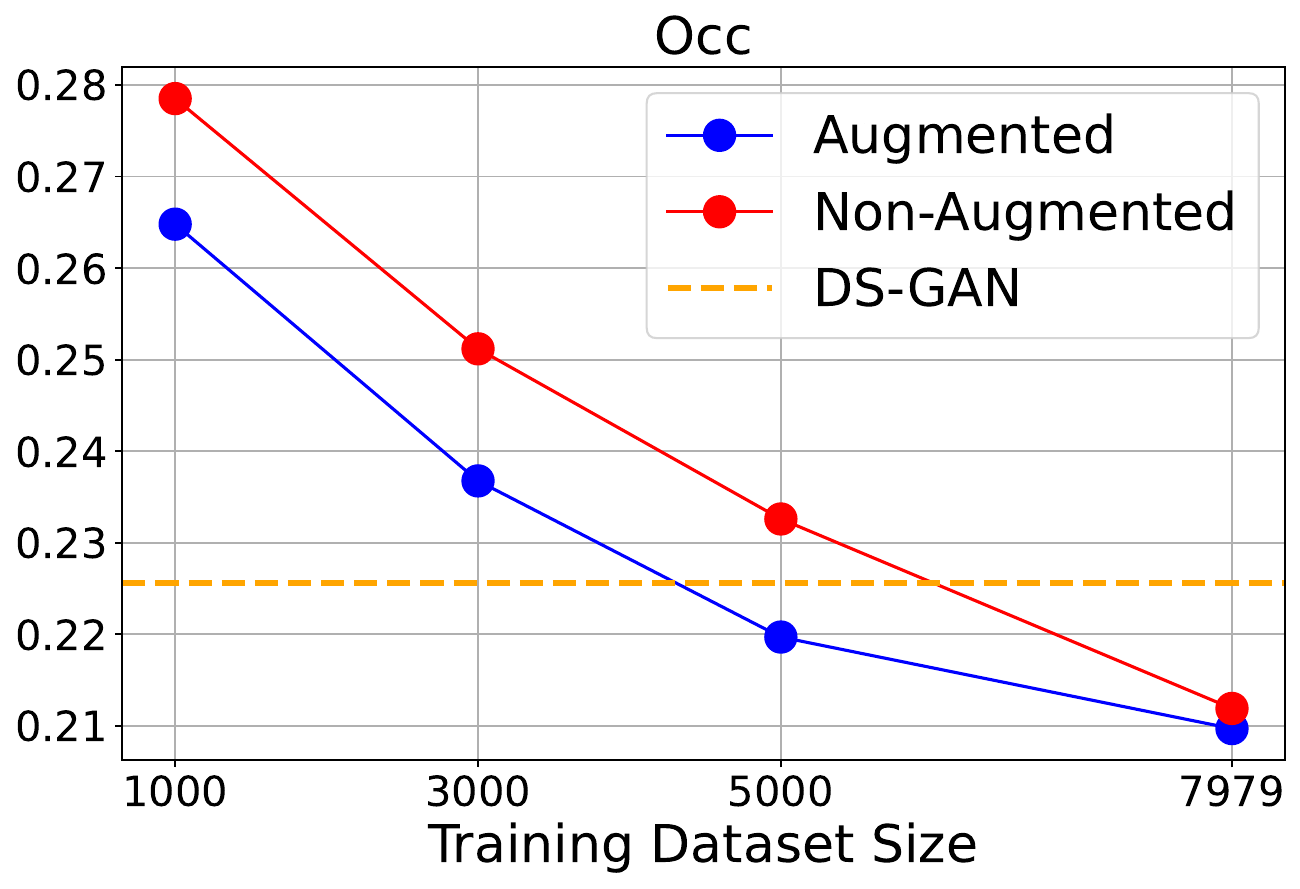}
        \end{subfigure}
         &
        \begin{subfigure}{0.48\textwidth}
            \centering
            \includegraphics[width=\textwidth]{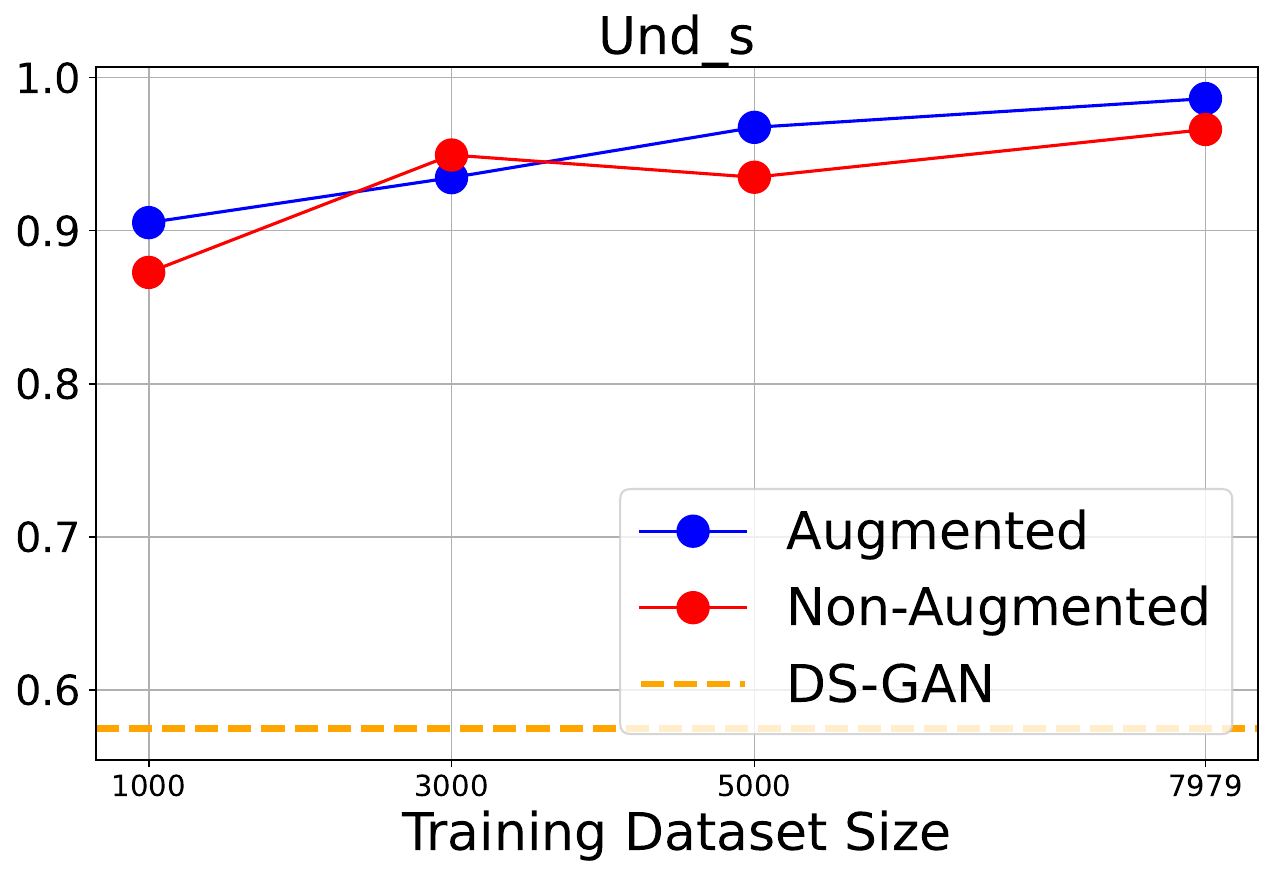}
        \end{subfigure}
    \end{tabular}
    \caption{Evaluation over the different training dataset sizes on PKU Dataset}
    \label{fig:Effect of Data Size}
\end{figure}

Here, we demonstrate the efficiency of utilizing the language model in a data scarcity setting. We train our model on the PKU dataset with different dataset sizes and evaluated its performance. \Fref{fig:Effect of Data Size} shows these results, particularly highlighting that the model can generate high-quality layouts even in small data settings like the 1k dataset. Additionally, from these results, we can infer the potential causes of the small degradation in \texttt{FD} and \texttt{und\_s} on the CGL dataset mentioned in the manuscript Effect of Augmentation Analysis.
The possible reason could be \texttt{FD}'s sensitivity to the small number of the test samples; Calculating \texttt{FD} with 3000 samples can introduce some variance. Another factor might be the varying effect of augmentation across different training data sizes. The gap of \texttt{FD} and \texttt{und\_s} across different dataset sizes in \Fref{fig:Effect of Data Size} diminishes increasing the data scale. We speculate that this reduction in performance differences introduces variance. Additionally, while \texttt{und\_s} has slightly decreased with augmentation in the CGL dataset (0.9987 → 0.9905), it still surpasses the value of real data (0.9884).
\section{Code Renderer}
The utilization of LLM in layout generation in this study differs from the approach of learning numerical information from bounding boxes. Instead of learning the numerical values associated with bounding boxes, the generated layouts are processed in the form of strings. This approach introduces the possibility of encountering two issues during the rendering process of the generated HTML code: abnormal formats and bounding box overflow beyond the constraints of the canvas. To assess the impact of these potential problems, we evaluate the test set, designating instances of such issues as failure cases. The evaluation of the test set reveals that there are no instances of failure cases. Consequently, the code renderer for Figure 1 in the manuscript is employed as a module that simply extracts and arranges string-formatted layout elements from HTML.
\section{Data Leckage Example}

\begin{figure}[tb]
  \centering
  \includegraphics[width=\linewidth]{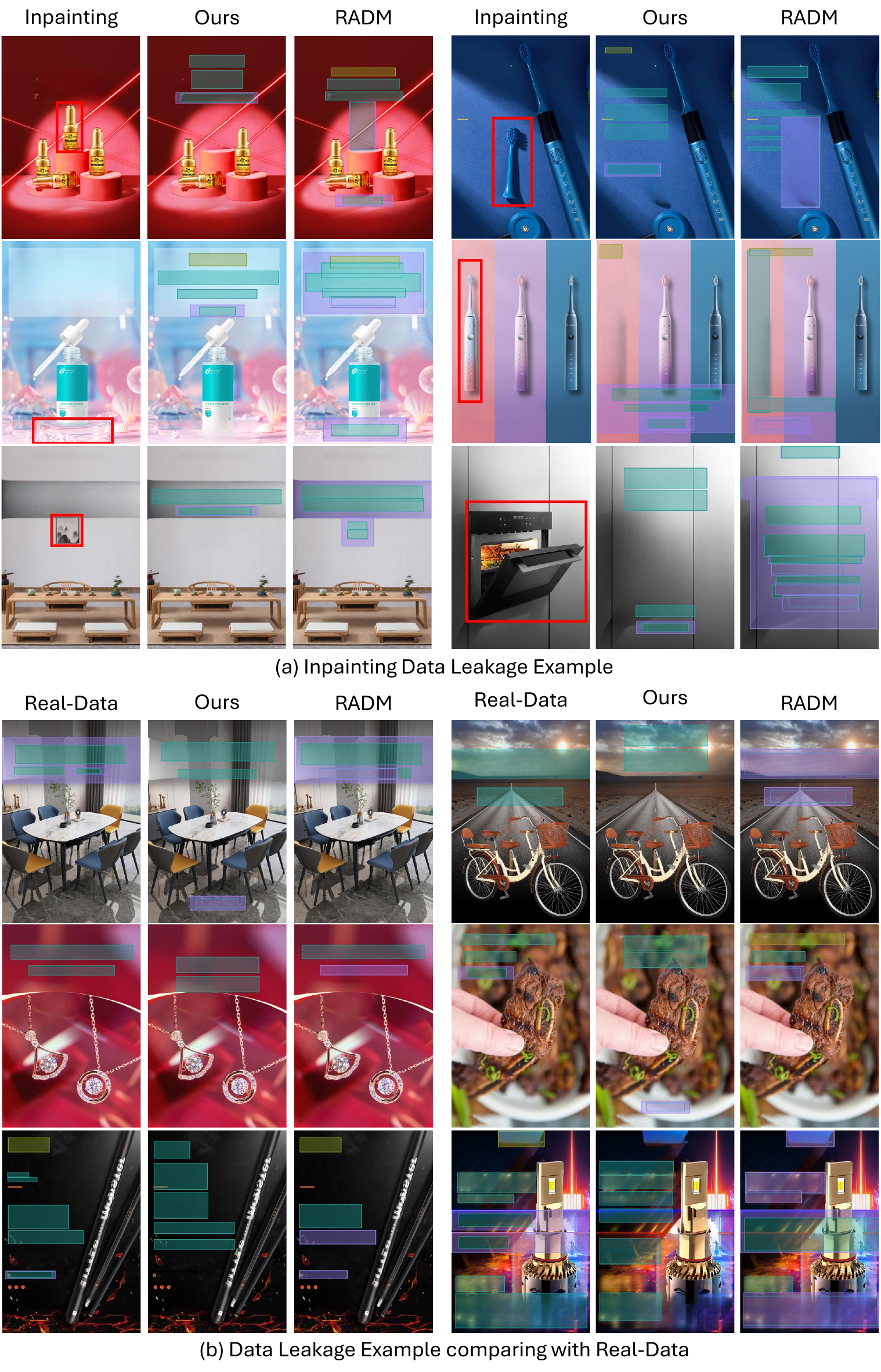}
  \caption{Visualization of Data Leakage Example.}
  \label{fig:Data_Leakage}
\end{figure}

We thoroughly examine the data leakage mentioned in the main text. We illustrate a more detailed example of data leakage in \Fref{fig:Data_Leakage}. (a) depicts an additional instance of data leakage, showcasing the generated layout within the inpainted area, as highlighted in the main paper. As we can observe, RADM consistently generates a layout at the inpainted location. (b) presents a comparison between the generated layouts of PosterLlama and RADM concerning annotated test data. The comparison demonstrates that the RADM's result closely resembles the structure of real data. These two results qualitatively verify that the inpainting area significantly influences \texttt{FD}, \texttt{occ}, and \texttt{rea}.
\section{Additional Qualitative Result}
In this section, we provide further qualitative findings of PosterLlama. These results demonstrate the effectiveness and versatility of our method across various layout generation scenarios.
\subsection{Undeterministic Sampling}
\begin{figure}[tb]
  \centering
  \includegraphics[width=\linewidth]{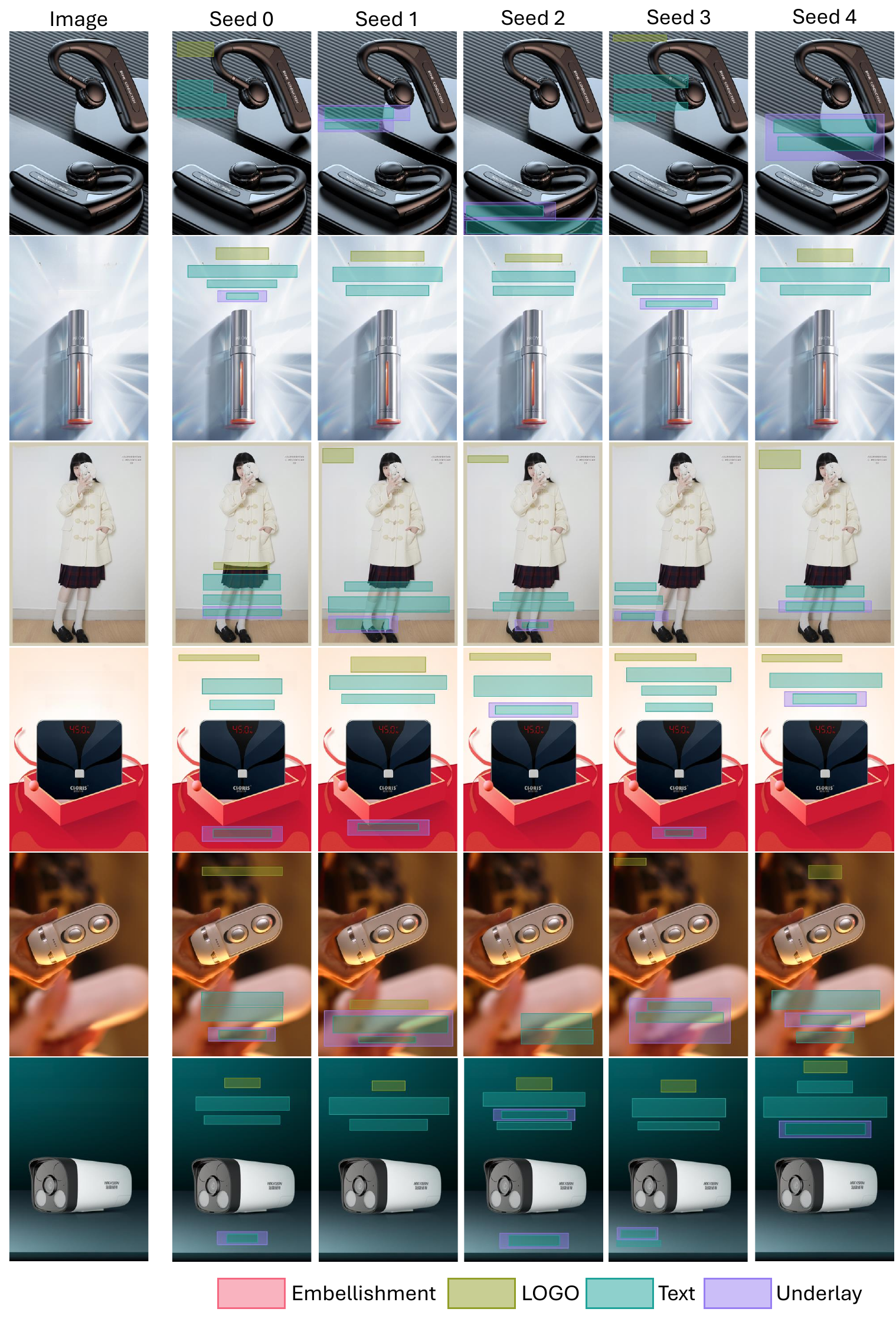}
  \caption{Visualization of image conditional layout generation across diverse seed.}
  \label{fig:Undeterministic}
\end{figure}

To verify our model's capability of creating diverse layouts as mentioned in the introduction of our paper, leveraging temperature and top-p \cite{holtzman2019curious} settings, we execute experiments with a text-agnostic model. We visualize the layouts generated from images with a range of salient objects about various seeds. \Fref{fig:Undeterministic} presents the outcomes. As observed, the results demonstrate that our PosterLlama can produce a variety of layouts while avoiding abnormal occlusions of salient objects and maintaining a high-quality, plausible layout.
\subsection{Diverse Conditional Generation}
We demonstrate PosterLlama's capabilities in conditional generation tasks, which are essential for real-world applications. Drawing from the approach of contents-agnostic layout generation as defined in previous work \cite{jiang2023layoutformer++}, we expand upon the main paper by adding two additional conditions including Gen-ITP and Refinement, resulting in a total of seven conditions for our model's conditional generation tasks.  Gen-ITP accepts an image along with element types and positions to generate size. The refinement task is aimed at correcting distorted layouts. For this refinement process, we introduce perturbations to the real layout using a normal distribution with a mean of 0 and a standard deviation of 0.01, following \cite{rahman2021ruite, jiang2023layoutformer++}. \Fref{fig:Conditional} and \ref{fig:Conditional2} shows the diverse conditional layout generation of PosterLlama including Gen-I, Gen-IT, Gen ITS, Gen-ITP, Recover, Completion and Refinement task about CGL-Dataset. The result demonstrates that our PosterLlama successfully generates an aligned, rational layout even in a conditional generation. In particular, the refinement and complement in \Fref{fig:Conditional2} show impressive results about distorted elements.
\subsection{Augmentation Example}
In the supplementary material, we present additional results of depth-guided augmentation in Figure \ref{fig:Augmentation}. The red boxes in the two samples below highlight areas affected by inpainting artifacts. As observed, the inpainting artifacts are eliminated in the results of the augmentation. These outcomes demonstrate that our method does more than just simple data augmentation; it also effectively reduces the impact of inpainting artifacts.

\begin{figure}[tb]
  \centering
  \includegraphics[width=\linewidth]{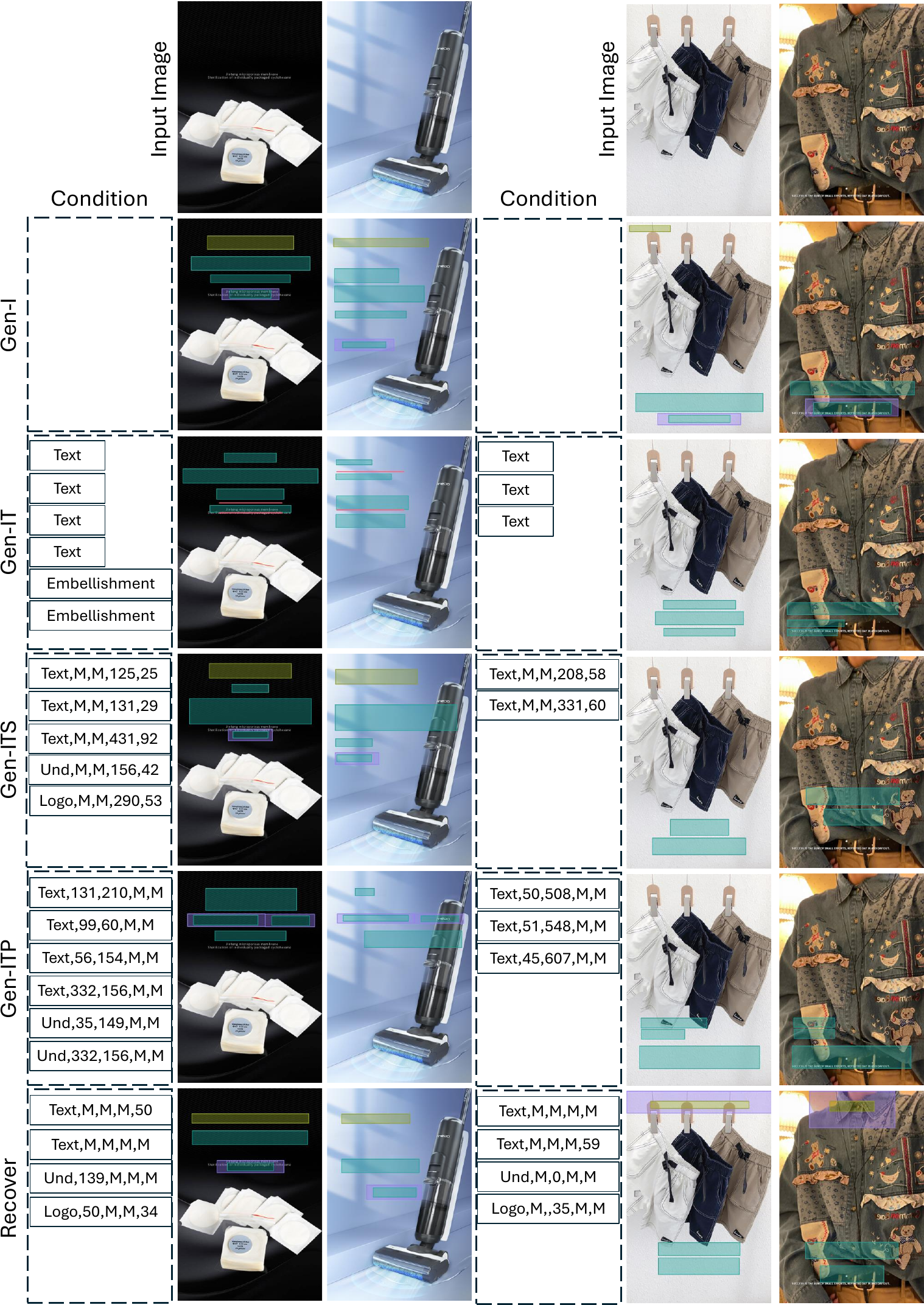}
  \caption{Visualization of 5 conditional generation tasks including Gen-I, Gen-IT, Gen-ITS, Gen-ITP, Recover.}
  \label{fig:Conditional}
\end{figure}

\begin{figure}[tb]
  \centering
  \includegraphics[width=\linewidth]{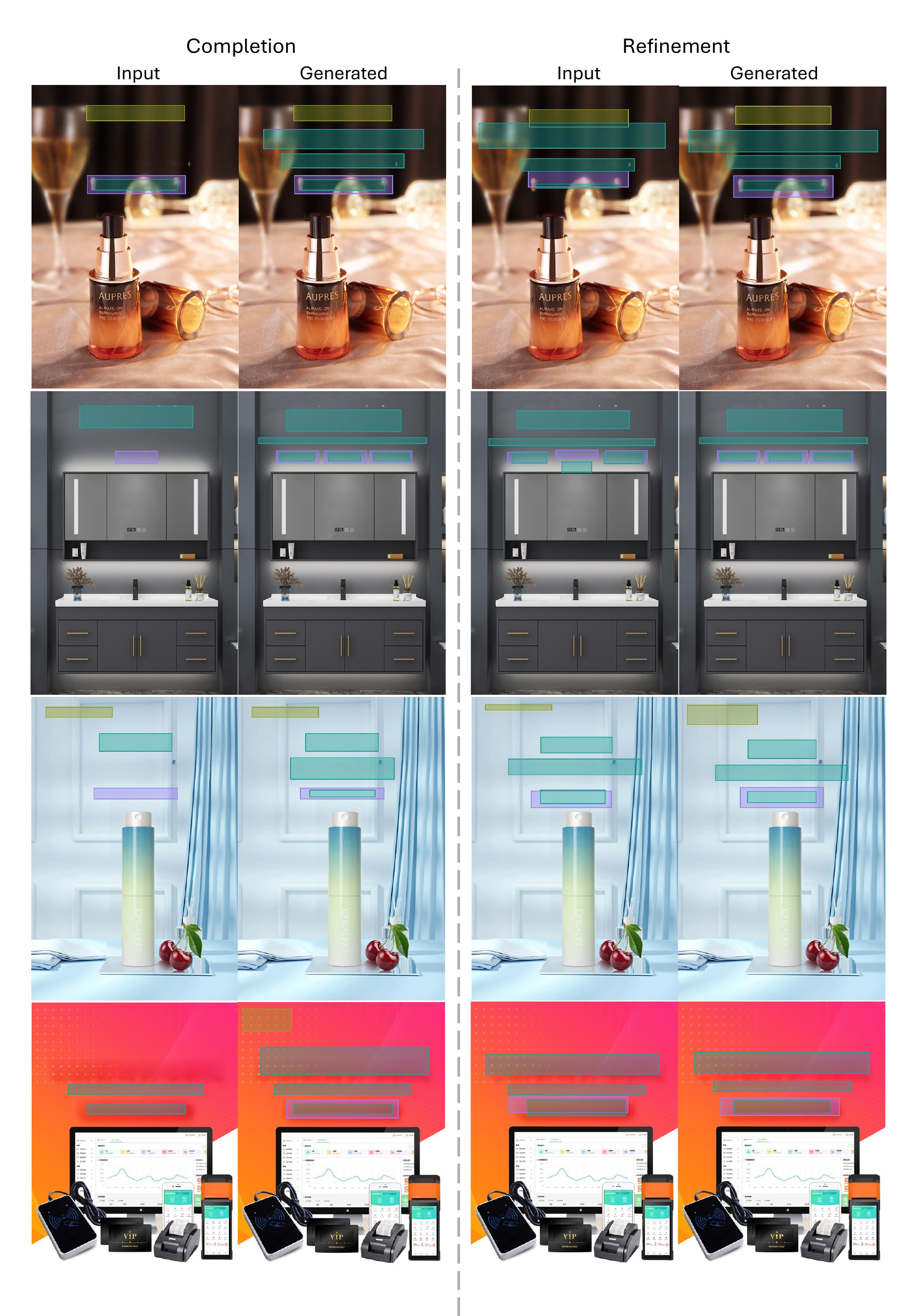}
  \caption{Visualization of 2 conditional generation including Completion and Refinement.}
  \label{fig:Conditional2}
\end{figure}

\begin{figure}[tb]
  \centering
  \includegraphics[width=\linewidth]{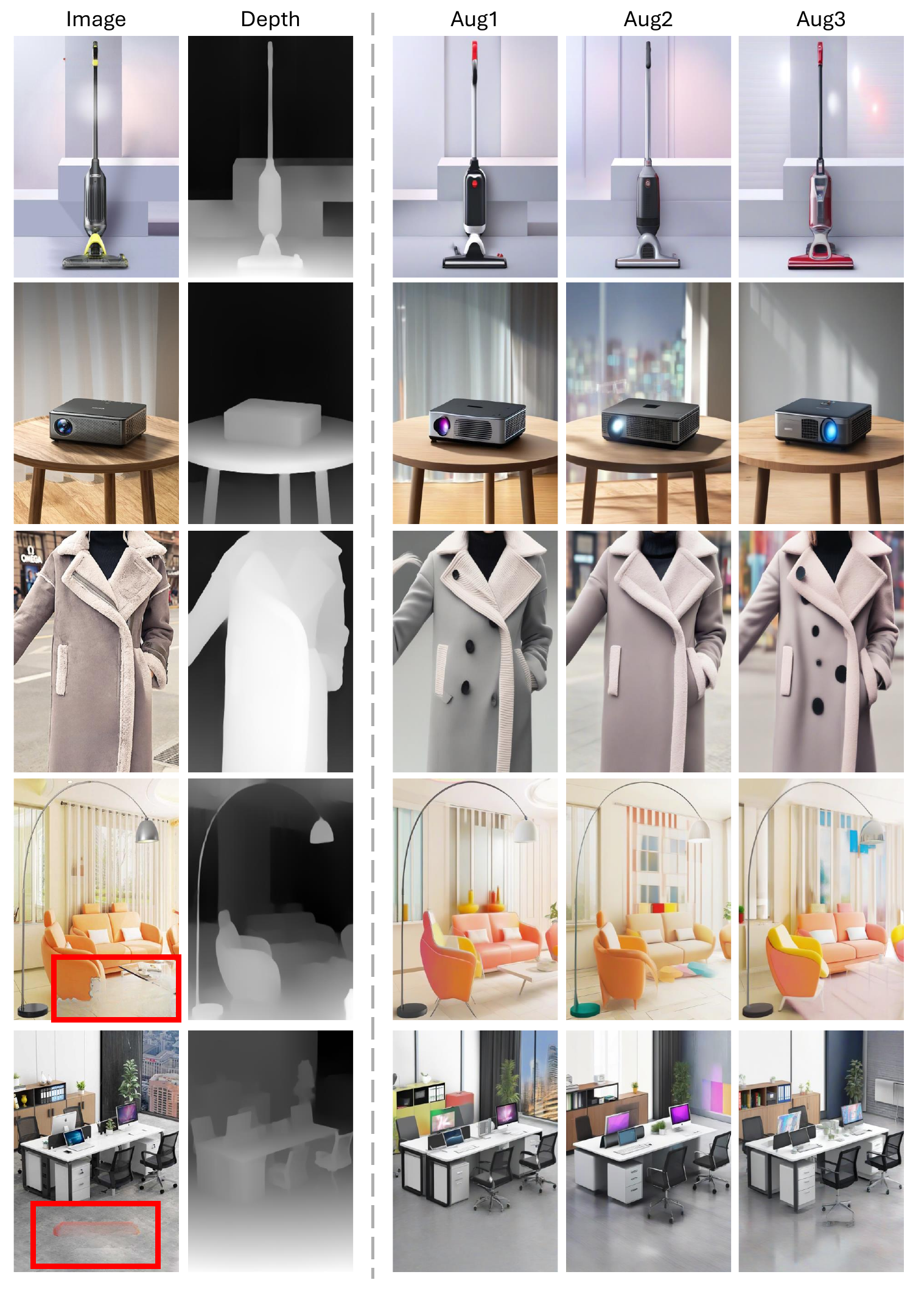}
  \caption{Visualization of depth-guided augmentation. Red-box is inpainting artifacts. }
  \label{fig:Augmentation}
\end{figure}
\clearpage
\newpage

\section{Input-Output Prompt Example}
Here, we present an example prompt for specifying a layout generation task. We illustrate a Recover constraint example that provides an overview of all tasks.
\vspace{10pt}
\begin{table}[h!]
\centering
\caption{Recover Prompt Example}
\begin{tabular}{p{\linewidth}}
\toprule
\textbf{Input} \\
    
\midrule
    \textbf{Task Definition}  \\ 
    I want to generate layout in poster design format. Please recover the layout html   according to the bbox , categories, size, image I provide (in html format)\\
    \\
    \textbf{Text Constraint}  \\
    Text :  ECCV2024 \& PosterLlama \& Try Now! \\
    \\
    \textbf{HTML Format}
\begin{lstlisting}[aboveskip=0pt, belowskip=0pt]
###bbox html:
<html> 
<body>  
<svg width = "513", height = "750"> `
<rect data-category="Text", x="172", y="<M>", width="<M>", height="29"/>
<rect data-category="Text", x="<M>", y="199", width="<M>", height="41"/>
<rect data-category="Text", x="<M>", y="201", width="162", height="39"/>
<rect data-category="Text", x="40", y="119", width="<M>", height="<M>"/>
<rect data-category="Underlay", x="190", y="16", width="<M>", height="<M>"/>
<rect data-category="Logo", x="<M>", y="189", width="<M>", height="64"/>
</svg> 
</body>
</html>\end{lstlisting}\\
\midrule
\textbf{Output} \\
\midrule
\begin{lstlisting}[aboveskip=0pt, belowskip=0pt]
<html> 
<body>  
<svg width = "513", height = "750">
<rect data-category="Text", x="172", y="80", width="179", height="29"/>
<rect data-category="Text", x="75", y="199", width="197", height="41"/>
<rect data-category="Text", x="282", y="201", width="162", height="39"/>
<rect data-category="Text", x="40", y="119", width="45", height="58"/>
<rect data-category="Underlay", x="190", y="16", width="149", height="61"/>
<rect data-category="Logo", x="55", y="189", width="408", height="64"/>
</svg> 
</body>
</html>\end{lstlisting} \\
\bottomrule
\end{tabular}
\end{table}

\clearpage

\end{document}